%% file: submit.tex
\documentclass[acmtog]{acmart}
\acmSubmissionID{1344}

\usepackage{booktabs} 

\usepackage{cancel}
\usepackage{pifont}
\newcommand{\xmark}{\ding{55}} 
\newcommand{\cmark}{\ding{51}} %

\usepackage[ruled]{algorithm2e} 

\SetAlFnt{\small}
\SetAlCapFnt{\small}
\SetAlCapNameFnt{\small}
\SetAlCapHSkip{0pt}

\acmJournal{TOG}

\acmYear{2025} 
\acmVolume{44} 
\acmNumber{4} 
\acmArticle{} 
\acmMonth{8}

\setcopyright{cc}
\setcctype{by-nc-nd}

\acmPrice{}
\acmDOI{10.1145/3731430}

\begin{document}
\title{FLoD: Integrating Flexible Level of Detail into 3D Gaussian Splatting for Customizable Rendering}

\author{Yunji Seo}
\authornote{Joint first authors; equal contribution.}
\orcid{0009-0004-9941-3610}
\affiliation{%
  \institution{Yonsei University}
  \country{South Korea}}
\email{oungji@yonsei.ac.kr}
\author{Young Sun Choi}
\authornotemark[1]
\orcid{0009-0001-9836-4245}
\affiliation{%
  \institution{Yonsei University}
  \country{South Korea}
}
\email{youngsun.choi@yonsei.ac.kr}
\author{HyunSeung Son}
\orcid{0009-0009-1239-0492}
\affiliation{
    \institution{Yonsei University}
    \country{South Korea}
}
\email{ghfod0917@yonsei.ac.kr}
\author{Youngjung Uh}
\authornote{Corresponding author.}
\orcid{0000-0001-8173-3334}
\affiliation{
    \institution{Yonsei University}
    \country{South Korea}
}
\email{yj.uh@yonsei.ac.kr}

\begin{teaserfigure}
  \includegraphics[width=\textwidth]{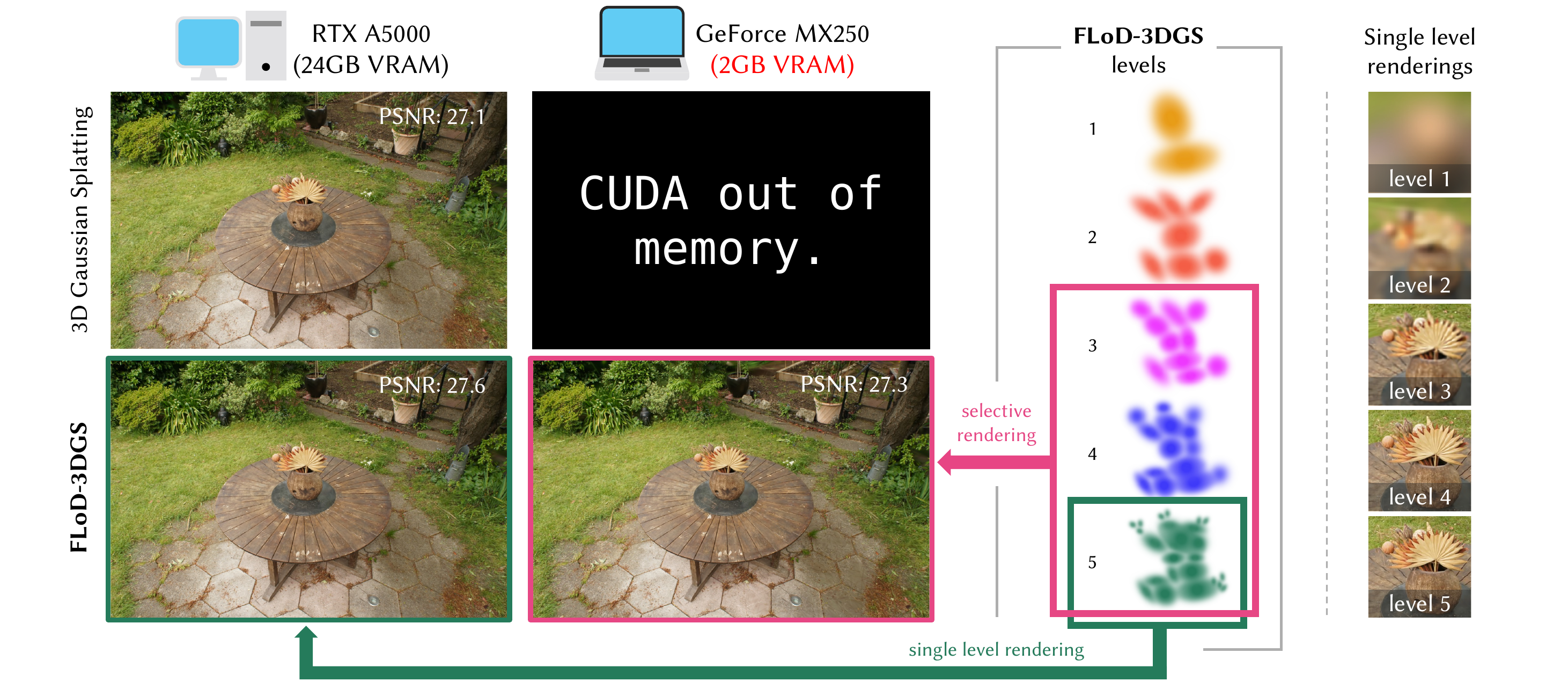}
    \caption{
    We introduce Level of Detail (LoD) mechanism in  3D Gaussian Splatting (3DGS) through multi-level representations. These representations enable flexible rendering by selecting individual levels or subsets of levels. The green box illustrates max-level rendering on a high-end server, while the pink box shows subset-level rendering for a low-cost laptop, where traditional 3DGS fails to render. Thus, FLoD-3DGS can flexibly adapt to diverse hardware settings.
    }  \Description{}
  \label{fig:teaser}
\end{teaserfigure}

\begin{abstract}

3D Gaussian Splatting (3DGS) has significantly advanced computer graphics by enabling high-quality 3D reconstruction and fast rendering speeds, inspiring numerous follow-up studies. However, 3DGS and its subsequent works are restricted to specific hardware setups, either on only low-cost or on only high-end configurations. Approaches aimed at reducing 3DGS memory usage enable rendering on low-cost GPU but compromise rendering quality, which fails to leverage the hardware capabilities in the case of higher-end GPU. Conversely, methods that enhance rendering quality require high-end GPU with large VRAM, making such methods impractical for lower-end devices with limited memory capacity. Consequently, 3DGS-based works generally assume a single hardware setup and lack the flexibility to adapt to varying hardware constraints. 

To overcome this limitation, we propose Flexible Level of Detail (FLoD) for 3DGS. FLoD constructs a multi-level 3DGS representation through level-specific 3D scale constraints, where each level independently reconstructs the entire scene with varying detail and GPU memory usage. A level-by-level training strategy is introduced to ensure structural consistency across levels. Furthermore, the multi-level structure of FLoD allows selective rendering of image regions at different detail levels, providing additional memory-efficient rendering options. To our knowledge, among prior works which incorporate the concept of Level of Detail (LoD) with 3DGS, FLoD is the first to follow the core principle of LoD by offering adjustable options for a broad range of GPU settings.

Experiments demonstrate that FLoD provides various rendering options with trade-offs between quality and memory usage, enabling real-time rendering under diverse memory constraints. Furthermore, we show that FLoD generalizes to different 3DGS frameworks, indicating its potential for integration into future state-of-the-art developments.

\end{abstract}

%
%

%
%


\maketitle

\input{main}

\bibliographystyle{ACM-Reference-Format}
\bibliography{ourbib}

\input{supple}

\end{document}

%% file: main.tex
\section{Introduction}

Recent advances in 3D reconstruction have led to significant improvements in the fidelity and rendering speed of novel view synthesis. 
In particular, 3D Gaussian Splatting (3DGS)\cite{kerbl3Dgaussians} has demonstrated photo-realistic quality at exceptionally fast rendering rates. However, its reliance on numerous Gaussian primitives makes it impractical for rendering on devices with limited GPU memory. Similarly, methods such as AbsGS\cite{ye2024absgs}, FreGS~\cite{zhang2024fregs}, and Mip-Splatting~\cite{yu2024mipsplatting}, which further enhance rendering quality, remain constrained to higher-end devices due to their dependence on a comparable or even greater number of Gaussians for scene reconstruction.
Conversely, LightGaussian~\cite{fan2023lightgaussian} and CompactGS~\cite{lee2023compact} address memory limitations by removing redundant Gaussians, which helps reduce rendering memory demands as well as reducing storage size.
However, the reduction in memory usage comes at the expense of rendering quality.
Consequently, existing approaches are developed based on either high-end or low-cost devices. As a result, they lack the flexibility to adapt and produce optimal renderings across various GPU memory capacities.

Motivated by the need for greater flexibility, we integrate the concept of Level of Detail (LoD) within the 3DGS framework.
LoD is a concept in graphics and 3D modeling that provides different levels of detail, allowing model complexity to be adjusted for optimal performance on varying devices.
At lower levels, models possess reduced geometric and textural detail, which decreases memory and computational demands. 
Conversely, at higher levels, models have increased detail, leading to higher memory and computational demands. 
This approach enables graphical applications to operate effectively on systems with varying GPU settings, avoiding processing delays for low-end devices while maximizing visual quality for high-end setups. 
Additionally, it enables the selective application of different levels, using higher levels where necessary and lower levels in less critical regions, to enhance resource efficiency while maintaining a high perceptual image.

Recent methods that integrate LoD with 3DGS~\cite{ren2024octreegs, kerbl2024hierarchicalgaussians, liu2024citygaussian} develop multi-level representations to achieve consistent and high-quality renderings, rather than the adaptability to diverse GPU memory settings.
While these methods excel at creating detailed high-level representations, rendering with only lower-level representations to accommodate middle or low-cost GPU settings causes significant scene content loss and distortions.
This highlights the lack of flexibility in existing methods to adapt and optimize rendering quality across different hardware setups.

\begin{figure*}[ht]
    \includegraphics[width=\textwidth]{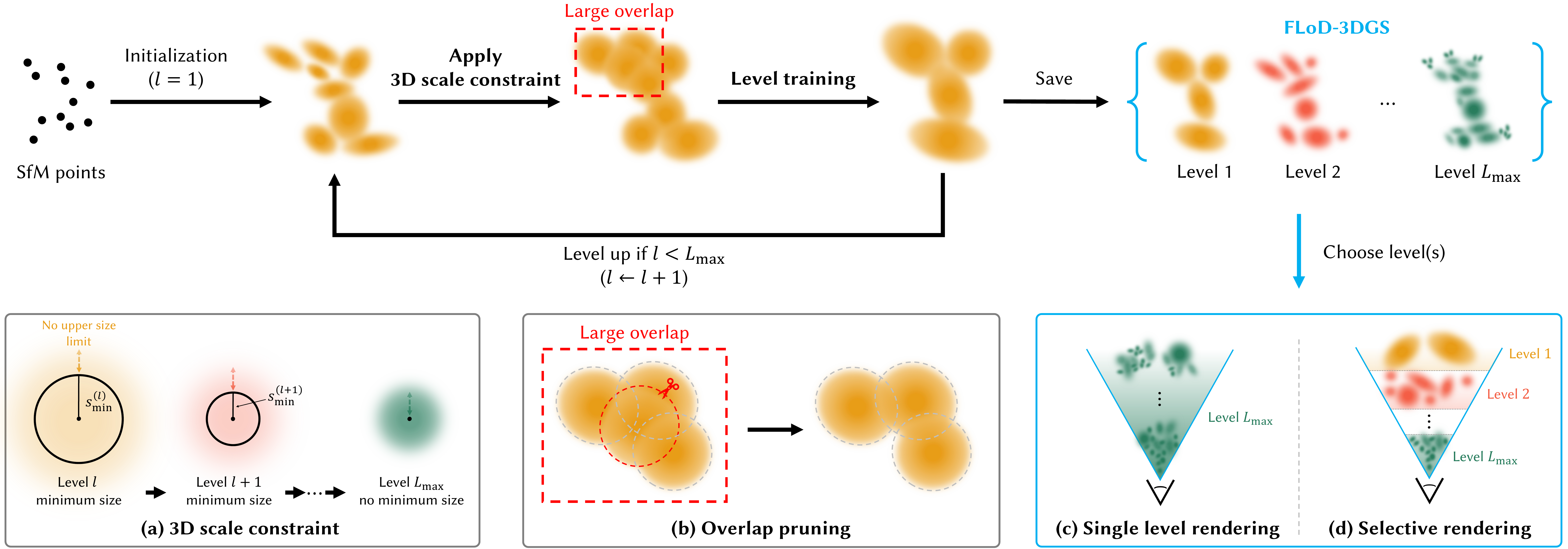}
    \caption{Method overview. Training begins at level 1, initialized from SfM points. During the training of each level, (a) a level-specific 3D scale constraint \(s_\text{min}^{(l)}\) is imposed on the Gaussians as a lower bound, and (b) overlap pruning is performed to mitigate Gaussian overlap. At the end of each level’s training, the Gaussians are cloned and saved as the final representation for level \(l\). This level-by-level training continues until the max level (\(L_\text{max}\)), resulting in a multi-level 3D Gaussian representation referred to as FLoD-3DGS. FLoD-3DGS supports (c) single-level rendering and (d) selective rendering using multiple levels.} \Description{}
    \label{fig:method} 
\end{figure*}

To address the hardware adaptability challenges, we propose Flexible Level of Detail (FLoD).
FLoD constructs a multi-level 3D Gaussian Splatting (3DGS) representation that provides varying levels of detail and memory requirements, with each level independently capable of reconstructing the full scene.
Our method applies a level-specific 3D scale constraint, which increases each successive level, to limit the amount of detail reconstructed and the rendering memory demand. 
Furthermore, we introduce a level-by-level training method to maintain a consistent 3D structure across all levels.
Our trained FLoD representation provides the flexibility to choose any single level based on the available GPU memory or desired rendering rates. 
Furthermore, the independent and multi-level structure of our method allows different parts of an image to be rendered with different levels of detail, which we refer to as selective rendering. 
Depending on the scene type or the object of interest, higher-level Gaussians can be used to rasterize important regions, while lower levels can be assigned to less critical areas, resulting in more efficient rendering.
As a result, FLoD provides the versatility of adapting to diverse GPU settings and rendering contexts.

We empirically validate the effectiveness of FLoD in offering flexible rendering options, tested on both a high-end server and a low-cost laptop.
We conduct experiments not only on the Tanks and Temples~\cite{Knapitsch2017tandt} and Mip-Nerf360~\cite{barron2022mipnerf360} datasets, which are commonly used in 3DGS and its variants but also on the DL3DV-10K~\cite{ling2023dl3dv10k} dataset, which contains distant background elements that can be effectively represented through LoD.
Furthermore, we demonstrate that FLoD can be easily integrated into existing 3DGS variants, while also enhancing the rendering quality.

\section{Related Work}

\subsection{3D Gaussian Splatting}
3D Gaussian Splatting (3DGS)~\cite{kerbl3Dgaussians} has attained popularity for its fast rendering speed in comparison to other novel view synthesis literature such as NeRF~\cite{mildenhall2020nerf}.
Subsequent works, such as FreGS~\cite{zhang2024fregs} and AbsGS~\cite{ye2024absgs}, improve rendering quality by modifying the loss function and the Gaussian density control strategy, respectively.
However, these methods, including 3DGS, demand high rendering memory because they rely on a large number of Gaussians, making them unsuitable for low-cost devices with limited GPU memory.

To address these memory challenges, various works have proposed compression methods for 3DGS.
LightGaussian~\cite{fan2023lightgaussian} and Compact3D~\cite{lee2023compact} use pruning techniques, while EAGLES~\cite{girish2024eagles} employs quantized embeddings.
However, their rendering quality falls short compared to 3DGS. 
RadSplat~\cite{niemeyer2024radsplat} and Scaffold-GS~\cite{lu2023scaffoldgs} maintain rendering quality while reducing memory usage with neural radiance field prior and neural Gaussians.
Despite these advancements, existing 3DGS methods lack the flexibility to provide multiple rendering options for optimizing performance across various GPU settings.

In contrast, we propose a multi-level 3DGS that increases rendering flexibility by enabling rendering across various GPU settings, ranging from server GPUs with 24GB VRAM to laptop GPUs with 2GB VRAM.

\subsection{Multi-Scale Representation}
There have been various attempts to improve the rendering quality of novel view synthesis through multi-scale representations. 
In the field of Neural Radiance Fields (NeRF), approaches such as Mip-NeRF~\cite{barron2021mipnerf} and Zip-NeRF~\cite{barron2023zipnerf} adopt multi-scale representations to improve rendering fidelity.
Similarly, in 3D Gaussian Splatting (3DGS), Mip-Splatting~\cite{yu2024mipsplatting} uses a multi-scale filtering mechanism, and MS-GS~\cite{yan2024ms-gs} applies a multi-scale aggregation strategy.
However, these methods primarily focus on addressing the aliasing problem and do not consider the flexibility to adapt to different GPU settings.

In contrast, our proposed method generates a multi-level representation that not only provides flexible rendering across various GPU settings but also enhances reconstruction accuracy.

\subsection{Level of Detail}
Level of Detail (LoD) in computer graphics traditionally uses multiple representations of varying complexity, allowing the selection of detail levels according to computational resources. 
In NeRF literature, NGLOD~\cite{takikawa2021nglod} and Variable Bitrate Neural Fields~\cite{takikawa2022vqad} create LoD structures based on grid-based NeRFs.

In 3D Gaussian Splatting (3DGS), methods such as Octree-GS~\cite{ren2024octreegs} and Hierarchical-3DGS~\cite{kerbl2024hierarchicalgaussians} integrate the concept of LoD and create multi-level 3DGS representation for efficient and high-detail rendering.
However, these methods primarily target efficient rendering on high-end GPUs, such as A6000 or A100 GPUs with 48GB or 80GB VRAM.
Moreover, these methods render using Gaussians from the entire range of levels, not solely from individual levels. Rendering with individual levels, particularly the lower ones, leads to a loss of image quality.
Therefore, theses methods cannot provide rendering options with lower memory demands.
While CityGaussian~\cite{liu2024citygaussian} can render individual levels using its multi-level representations created with various compression rates, it also does not address the challenges of rendering on lower-cost GPU.

In contrast, our method allows for rendering using either individual or multiple levels, as all levels independently reconstruct the scene.
Additionally, as each level has an appropriate degree of detail and corresponding rendering computational demand, our method offers rendering options that can be optimized for diverse GPU setups. 

\section{Preliminary}
\label{sec:prem}
3D Gaussian Splatting (3DGS) \cite{kerbl3Dgaussians} introduces a method to represent a 3D scene using a set of 3D Gaussian primitives.
Each 3D Gaussian is characterized by attributes: position \(\boldsymbol{\mu}\), opacity \(o\), covariance matrix \(\boldsymbol{\Sigma}\), and spherical harmonic coefficients.
The covariance matrix \(\mathbf{\Sigma}\) is factorized into a scaling matrix \(\mathbf{S}\) and a rotation matrix \(\mathbf{R}\):
\begin{equation}
\boldsymbol{\Sigma} = \mathbf{R} \mathbf{S} \mathbf{S}^\top \mathbf{R}^\top.
\end{equation}

To facilitate the independent optimization of both components, the scaling matrix \(\mathbf{S}\) is optimized through the vector \(\mathbf{s}_\text{opt}\), and the rotation matrix \(\mathbf{R}\) is optimized via the quaternion \(\mathbf{q}\).
These 3D Gaussians are projected to 2D screenspace and the opacity contribution of a Gaussian at a pixel \((x, y)\) is computed as follows:

\begin{equation}
\alpha(x,y) = o \cdot e^{-\frac{1}{2} \left( ([x,y]^T - \boldsymbol{\mu}')^T \boldsymbol{\Sigma}'^{-1} ([x,y]^T - \boldsymbol{\mu}') \right)},
\end{equation}

where \(\boldsymbol{\mu}'\) and \(\boldsymbol{\Sigma}'\) are the 2D projected mean and covariance matrix of the 3D Gaussians.
The image is rendered by alpha blending the projected Gaussians in depth order.

\section{Method: Flexible Level of Detail}
\label{sec:method}
Our method reconstructs a scene as a \(L_\text{max}\)-level 3D Gaussian representation, using 3D Gaussians of varying sizes from level 1 to \(L_\text{max}\) (Section~\ref{sec:scale_constraint}).
Through our level-by-level training process (Section~\ref{sec:level-by-level}), each level independently captures the overall scene structure while optimizing for render quality appropriate to its respective level.
This process results yields a novel LoD structure of 3D Gaussians, which we refer to as FLoD-3DGS. 
The lower levels in FLoD-3DGS reconstruct the coarse structures of the scene using fewer and larger Gaussians, while higher levels capture fine details using more and smaller Gaussians.
Additionally, we introduce overlap pruning to eliminate artifacts caused by excessive Gaussian overlap (Section~\ref{sec:overlap_prune}) and demonstrate our method’s easy integration with different 3DGS-based method (Section~\ref{sec:compatibility}).

\subsection{3D Scale Constraint}
\label{sec:scale_constraint}
For each level \(l\) where $l\in[1, L_\text{max}]$, we impose a 3D scale constraint \(s_\text{min}^{(l)}\) as the lower bound on 3D Gaussians. 
The 3D scale constraint \(s_\text{min}^{(l)}\) is defined as follows:
\begin{equation}
s_\text{min}^{(l)}=
\begin{cases}
\lambda \times \rho^{1-l} & \text{for } 1 \le l < L_\text{max}  \\
0 & \text{for } l = L_\text{max}.
\end{cases}
\end{equation}
\(\lambda\) is the initial 3D scale constraint, and \(\rho\) is the scale factor by which the 3D scale constraint is reduced for each subsequent level. 
The 3D scale constraint is 0 at \(L_\text{max}\) to allow reconstruction of the finest details without constraints at this stage.
Then, we define 3D Gaussians' scale at level $l$ as follows:
\begin{equation}
\label{eq:apply_scale_constraint}
\mathbf{s}^{(l)} = e^{\mathbf{s_\text{opt}}} + s_\text{min}^{(l)}.
\end{equation}
where \(\mathbf{s_\text{opt}}\) is the learnable parameter for scale, while the 3D scale constraint \(s_\text{min}^{(l)}\) is fixed.
We note that $\mathbf{s}^{(l)} >= s_\text{min}^{(l)}$ because $e^{\mathbf{s_\text{opt}}} > 0$.

On the other hand, there is no upper bound on Gaussian size at any level. 
This allows for flexible modeling, where scene contents with simple shapes and appearances can be modeled with fewer and larger Gaussians, avoiding the redundancy of using many small Gaussians at high levels. 

\subsection{Level-by-level Training}
\label{sec:level-by-level}
We design a coarse-to-fine training process, 
where the next-level Gaussians are initialized by the fully-trained previous-level Gaussians.  
Similar to 3DGS, the 3D Gaussians at level 1 are initialized from SFM points. 
Then, the training process begins. Note that training of subsequent levels are nearly identical.

The training process consists of periodic densification and pruning of Gaussians over a set number of iterations. This is then followed by the optimization of Gaussian attributes without any further densification or pruning for an additional set of iterations.
Throughout the entire training process for level \(l\), the 3D scale of the Gaussian is constrained to be larger or equal to \( s_\text{min}^{(l)} \) by definition.

After completing training at level \(l\), this stage is saved as a checkpoint. 
At this point, the Gaussians are cloned and saved as the final Gaussians for level \(l\). 
Then, the checkpoint Gaussians are used to initialize Gaussians of the next level \(l+1\).
For initialized Gaussians at the next level \(l+1\), we set
\begin{equation}
\label{eq:adjust_scale}
\mathbf{s}_\text{opt} = \textnormal{log}(\mathbf{s}^{(l)} - s_{\text{min}}^{(l+1)}),
\end{equation}
such that $\mathbf{s}^{(l+1)} = \mathbf{s}^{(l)}$.
It prevents abrupt initial loss by eliminating the gap $\mathbf{s}^{(l+1)} - \mathbf{s}^{(l)} = \cancel{e^{\mathbf{s_\text{opt}^\text{prev}}}} + s_\text{min}^{(l+1)} - (\cancel{e^{\mathbf{s_\text{opt}^\text{prev}}}} + s_\text{min}^{(l)})$. Note that \(\mathbf{s_\text{opt}^\text{prev}}\) represents the learnable parameter for scale at level \(l\). 

\subsection{Overlap Pruning}
\label{sec:overlap_prune}

To prevent rendering artifacts, we remove Gaussians with large overlaps.
Specifically, Gaussians whose average distance of its three nearest neighbors falls below a pre-defined distance threshold \( d_\text{OP}^{(l)} \) are eliminated.  
Equation for \( d_\text{avg}^{(l)} \) is given as:
\begin{equation}
d_\text{avg}^{(i)} = \frac{1}{3} \sum_{j=1}^{3} d_{ij}
\end{equation}
\( d_\text{OP}^{(l)} \) is set as half of the 3D scale constraint \( s_\text{min}^{(l)} \) for training level \( l \).
This method also reduces the overall memory footprint.

\subsection{Compatibility to Different Backbone}
\label{sec:compatibility}
The simplicity of our method, stemming from the straightforward design of the 3D scale constraints and the level-by-level training pipeline, makes it easy to integrate with other 3DGS-based techniques. 
We integrate our approach into Scaffold-GS \cite{lu2023scaffoldgs}, a variant of 3DGS that leverages anchor-based neural Gaussians.
We generate a multi-level set of Scaffold-GS by applying progressively decreasing 3D scale constraints on the neural Gaussians, optimized through our level-by-level training method.

\section{Rendering Methods}
FLoD's \(L_\text{max}\)-level 3D Gaussian representation provides a broad range of rendering options.
Users can select a single level to render the scene (Section~\ref{sec:single_rendering}), or multiple levels to increase rendering efficiency through selective rendering (Section~\ref{sec:selective_rendering}). 
Levels and rendering methods can be adjusted to achieve the desired rendering rates or to fit within available GPU memory limits.

\subsection{Single-level Rendering}
\label{sec:single_rendering}

From our multi-level set of 3D Gaussians \( \{\mathbf{G}^{(l)} \mid l = 1, \ldots, L_\text{max}\} \), users can choose any single level for rendering to match their GPU memory capabilities.
This approach is similar to how games or streaming services let users adjust quality settings to optimize performance for their devices.
Rendering any single level independently is possible because each level is designed to fully reconstruct the scene.

High-end hardware can handle the smaller and more numerous Gaussians of level \(L_\text{max}\), achieving high-quality rendering. However, rendering a large number of Gaussians may exceed the memory limits of commodity devices. In such cases, lower levels can be chosen to match the memory constraints.

\subsection{Selective Rendering}
\label{sec:selective_rendering}

\begin{figure}[ht]
    \includegraphics[width=0.94\columnwidth]{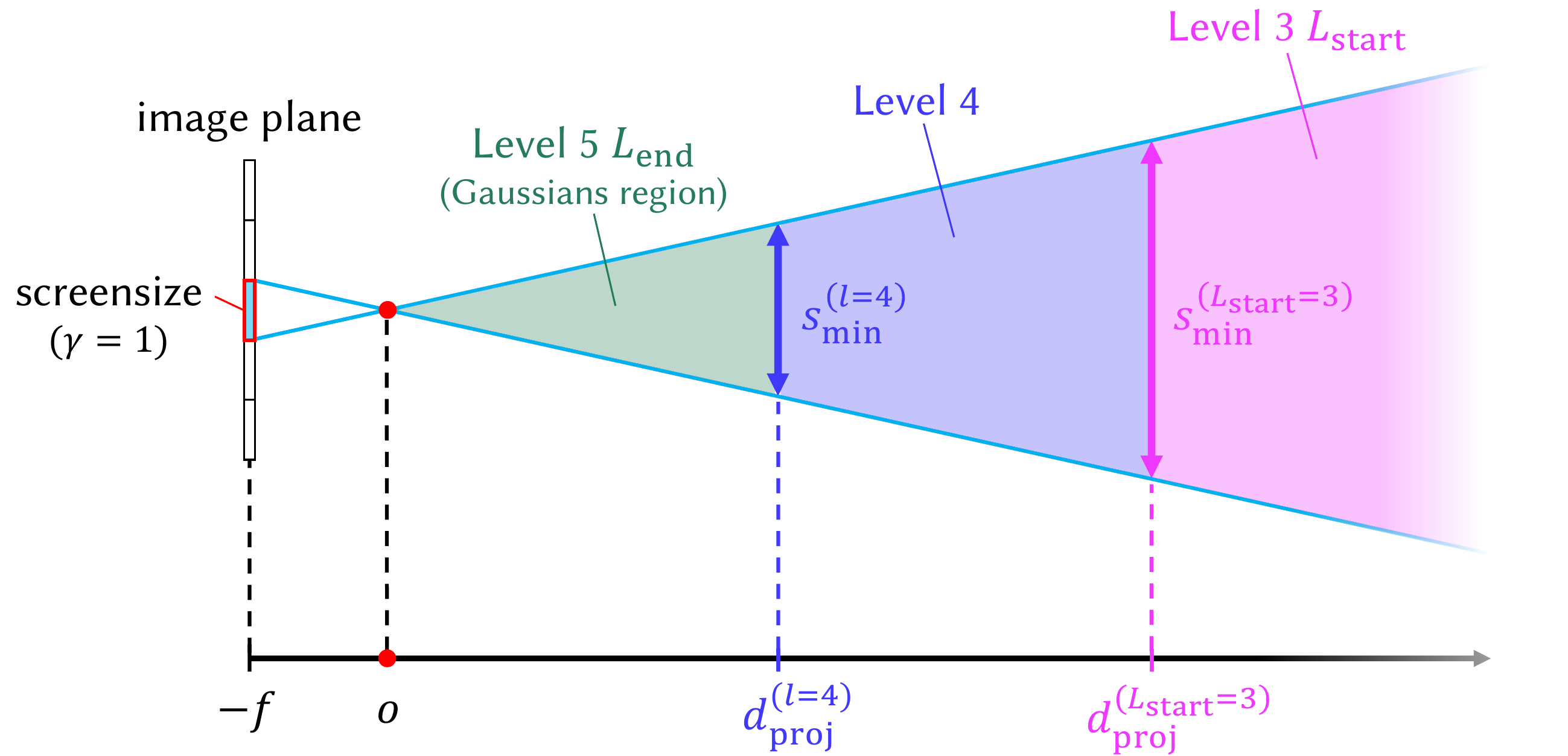}
    \caption{Visualization of the selective rendering process that shows how \(d_\text{proj}^{(l)}\) determines the appropriate Gaussian level for specific regions. This example visualizes the case where level 3 is used as \(L_\text{start}\) and level 5 as \(L_\text{end}\). } \Description{}
    \label{fig:rendering1}
\end{figure}

Although a single level can be simply selected to match GPU memory capabilities, utilizing multiple levels can further enhance visual quality while keeping memory demands manageable. 
Distant objects or background regions do not need to be rendered with high-level Gaussians, which capture small and intricate details. This is because the perceptual difference between high-level and low-level Gaussian reconstructions becomes less noticeable as the distance from the viewpoint increases.
In such scenarios, lower levels can be employed for distant regions while higher levels are used for closer areas. 
This arrangement of multiple level Gaussians can achieve perceptual quality comparable to using only high-level Gaussians but at a reduced memory cost. 

Therefore, we propose a faster and more memory-efficient rendering method by leveraging our multi-level set of 3D Gaussians \( \{\mathbf{G}^{(l)} \mid l = 1, \ldots, L_\text{max}\} \). 
We create the set of Gaussians \(\mathbf{G}_\text{sel}\) for selective rendering by sampling Gaussians from a desired level range, \(L_\text{start}\) to \(L_\text{end}\):
\begin{equation}
\mathbf{G}_\text{sel} = \bigcup_{l=L_\text{start}}^{L_\text{end}} \left\{ G^{(l)} \in \mathbf{G}^{(l)} \mid d_\text{proj}^{(l-1)} > d_{G^{(l)}} \geq d_\text{proj}^{(l)} \right\},
\end{equation}
where \(d_\text{proj}^{(l)}\) decides the inclusion of a Gaussian \(G^{(l)}\) whose distance from the camera is \(d_{G^{(l)}}\). 
We define \(d_\text{proj}^{(l)}\) as: 
\begin{equation}
d_\text{proj}^{(l)}= \frac{s_\text{min}^{(l)}}{\gamma} \times {f},
\label{eq:projected_size}
\end{equation}
by solving a proportional equation \(s_\text{min}^{(l)}:\gamma=d_\text{proj}^{(l)}:f\). Hence, the distance \(d_\text{proj}^{(l)}\) is where the level-specific Gaussian 3D scale constraint \(s_\text{min}^{(l)}\) becomes equal to the screen size threshold \(\gamma\) on the image plane. \(f\) is the focal length of the camera.
We set \(d_\text{proj}^{(L_\text{end})}=0\) and \(d_\text{proj}^{(L_\text{start} - 1)}=\infty\) to ensure that the scene is fully covered with Gaussians from the level range \(L_\text{start}\) to \(L_\text{end}\).

The Gaussian set \(\mathbf{G}_\text{sel}\) is created using the 3D scale constraint \(s_\text{min}^{(l)}\) because \(s_\text{min}^{(l)}\) represents the smallest 3D dimension that Gaussians at level \(l\) can be trained to represent. 
Therefore, the distance \(d_\text{proj}^{(l)}\) can be used to determine which level of Gaussians should be selected for different regions, as demonstrated in Figure~\ref{fig:rendering1}.
Since \(s_\text{min}^{(l)}\) is fixed for each level, \(d_\text{proj}^{(l)}\) is also fixed. 
Thus, constructing the Gaussian set \(\mathbf{G}_\text{sel}\) only requires calculating the distance of each Gaussian from the camera, \(d_{G^{(l)}}\). 
This method is computationally more efficient than the alternative, which requires calculating each Gaussian's 2D projection and comparing it with the screen size threshold \(\gamma\) at every level.

The threshold \(\gamma\) and the level range [\(L_\text{start}\), \(L_\text{end}\)] can be adjusted to accommodate specific memory limitations or desired rendering rates.
A smaller threshold and a high-level range prioritize fine details over memory and speed, while a larger threshold and a low-level range reduce memory use and speed up rendering at the cost of fine details.

\paragraph{Predetermined Gaussian Set}

\begin{figure}[t]
    \includegraphics[width=0.94\columnwidth]{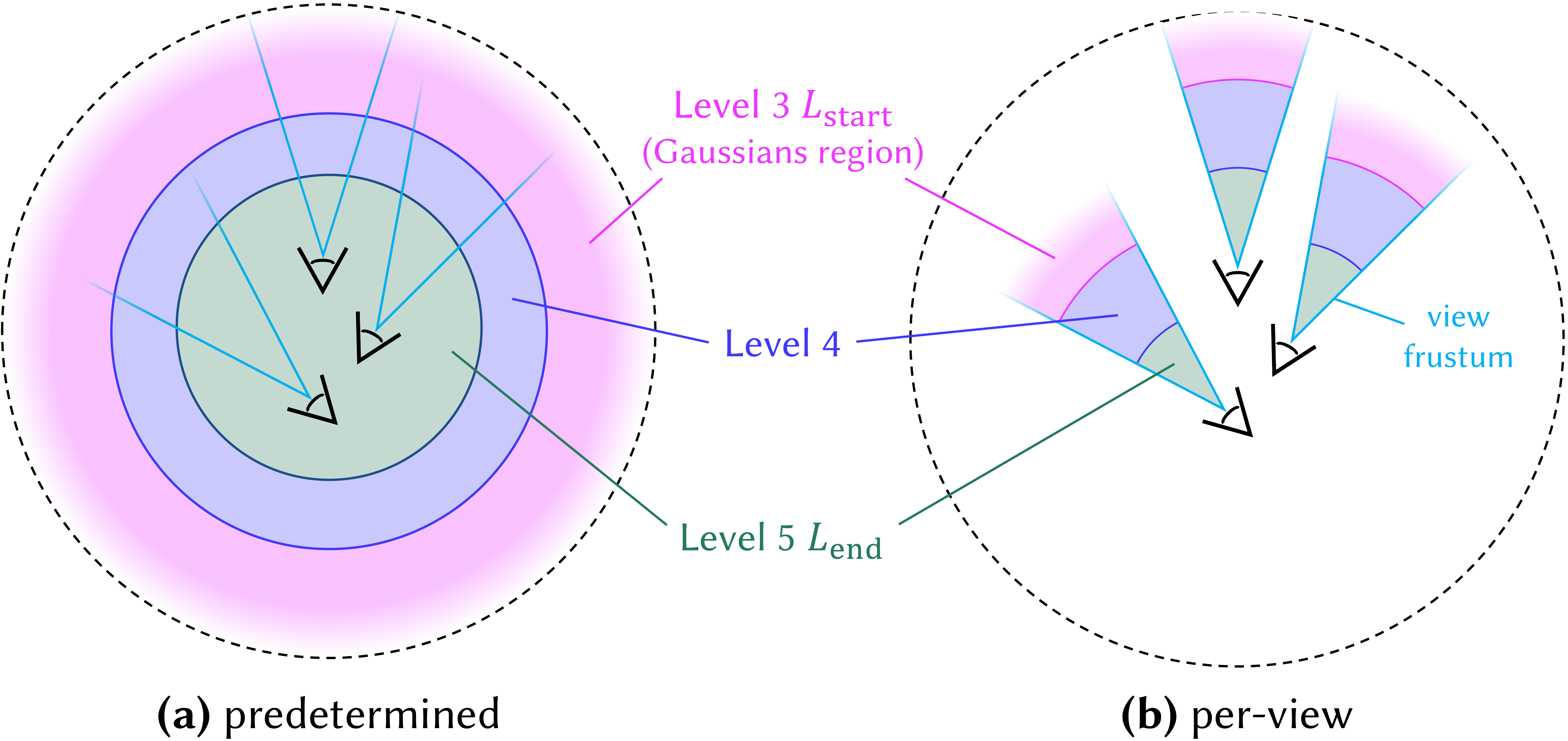}
    \caption{
    Comparison of predetermined Gaussian set \(\mathbf{G}_\text{sel}\) and per-view Gaussian set \(\mathbf{G}_\text{sel}\) creation methods. In the predetermined version, the Gaussian set is fixed, whereas the per-view version updates the Gaussian set dynamically whenever the camera position changes. This example illustrates the case where level 3 is used as \(L_\text{start}\) and level 5 as \(L_\text{end}\).
    }
    \label{fig:rendering2}
\end{figure}

For scenes where important objects are centrally located or the camera trajectory is confined to a small region, higher-level Gaussians can be assigned in the central areas, while lower-level Gaussians are allocated to the background.
This strategy enables high-quality rendering while reducing rendering memory and storage overhead.

To achieve this, we calculate the Gaussian distance \(d_{G^{(l)}}\) from the average position of all training view cameras before rendering and use it to predetermine the Gaussian subset \(\mathbf{G}_\text{sel}\), as illustrated in Figure~\ref{fig:rendering2}(a).
Since \(\mathbf{G}_\text{sel}\) is predetermined, it remains fixed during the rendering, eliminating the need to recalculate \(d_{G^{(l)}}\) whenever the camera view changes.
This predetermined approach allows for non-sampled Gaussians to be excluded, significantly reducing memory consumption during rendering. 
Furthermore, The sampled \(\mathbf{G}_\text{sel}\) can be stored for future use, requiring less storage compared to maintaining all level Gaussians.
As a result, this method is especially beneficial for low-cost devices with limited GPU memory and storage capacity.

\begin{figure*}[ht!]
    \includegraphics[width=\textwidth]{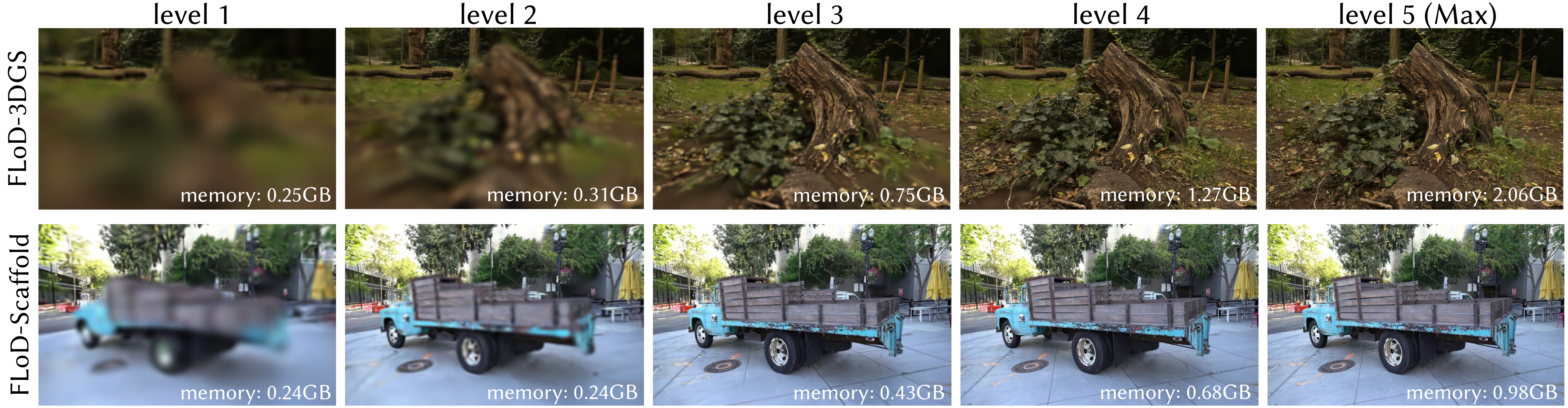}
    \caption{
    Renderings of each level in FLoD-3DGS and FLoD-Scaffold. FLoD can be integrated with both 3DGS and Scaffold-GS, with each level offering varying levels of detail and memory usage.
    }
    \label{fig:single}
\end{figure*}

\paragraph{Per-view Gaussian Set}
In large-scale scenes with camera trajectories that span broad regions, resampling the Gaussian set \(\mathbf{G}_\text{sel}\) based on the camera’s new position is necessary. 
This is because the camera may move and enter regions where lower level Gaussians have been assigned, leading to a noticeable decline in rendering quality.

Therefore, in such cases, we define the Gaussian distance \(d_{G^{(l)}}\) as the distance between a Gaussian \(G^{(l)}\) and the current camera position.
Consequently, whenever the camera position changes, \(d_{G^{(l)}}\) is recalculated to resample the Gaussian set \(\mathbf{G}_\text{sel}\) as illustrated in Figure~\ref{fig:rendering2}(b).
To maintain fast rendering rates, all Gaussians within the level range [\(L_\text{start}\), \(L_\text{end}\)] are kept in GPU memory.
Therefore, with the cost of increased rendering memory, selective rendering with per-view \(\mathbf{G}_\text{sel}\) effectively maintains consistent rendering quality over long camera trajectories. 


\section{Experiment}
\subsection{Experiment Settings}
\subsubsection{Datasets}
We conduct our experiments on a total of 15 real-world scenes. 
Two scenes are from Tanks\&Temples~\cite{Knapitsch2017tandt} and seven scenes are from Mip-NeRF360~\cite{barron2022mipnerf360}, encompassing both bounded and unbounded environments. 
These datasets are commonly used in existing 3DGS research. In addition, we incorporate six unbounded scenes from DL3DV-10K~\cite{ling2023dl3dv10k}, which include various urban and natural landscapes. 
We choose to include DL3DV-10K because it contains more objects located in distant backgrounds, providing a better demonstration of the diversity in real-world scenes.
Further details on the datasets can be found in Appendix~\ref{sec:appendix_dataset}. 

\subsubsection{Evaluation Metrics}
We measure PSNR, structural similarity SSIM~\cite{wang2004ssim}, and perceptual similarity LPIPS~\cite{zhang2018lpips} for a comprehensive evaluation. 
Additionally, we assess the number of Gaussians used for 
rendering the scenes, the GPU memory usage, and the rendering rates (FPS) to evaluate resource efficiency.

\subsubsection{Baselines}
We compare FLoD-3DGS against several models, including 3DGS~\cite{kerbl3Dgaussians}, Scaffold-GS~\cite{lu2023scaffoldgs}, Mip-Splatting\cite{yu2024mipsplatting}, Octree-GS~\cite{ren2024octreegs} and Hierarchical-3DGS\cite{kerbl2024hierarchicalgaussians}. 
Among these, the main competitors are Octree-GS and Hierarchical-3DGS, as they share the LoD concept with FLoD. 
However, these two competitors define individual level representation differently from ours.

In FLoD, each level representation independently reconstructs the scene.
In contrast, Octree-GS defines levels by aggregating the representations from the first level up to the specified level, meaning that individual levels do not exist independently. 
On the other hand, Hierarchical-3DGS does not have the concept of rendering using a specific level's representation, unlike FLoD and Octree-GS.
Instead, it employs a hierarchical structure with multiple levels, where Gaussians from different levels are selected based on the target granularity \(\tau\) setting for each camera view during rendering.

Additionally, like FLoD, Octree-GS is adaptable to both 3DGS and Scaffold-GS. 
We will refer to the 3DGS based Octree-GS as Octree-3DGS and the Scaffold-GS based Octree-GS as Octree-Scaffold.

\begin{figure*}[ht!]
    \includegraphics[width=\textwidth]{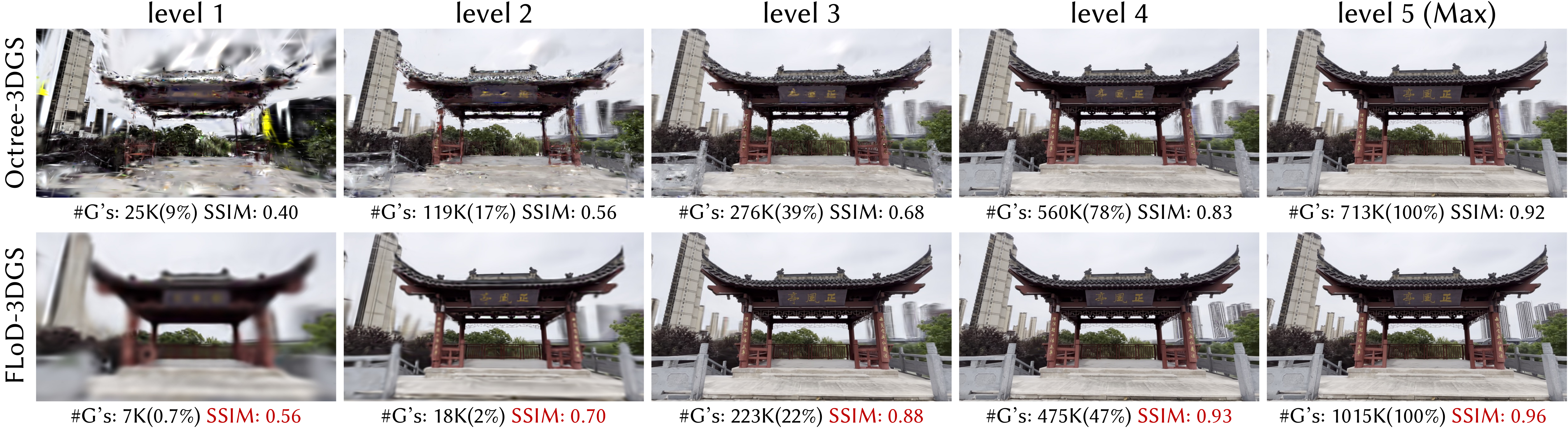}
    \caption{
    Comparison of the renderings at each level between FLoD-3DGS and Octree-3DGS on the DL3DV-10K dataset. 
    "\#G's" refers to the number of Gaussians, and the percentages (\%) next to these values indicate the proportion of Gaussians used relative to the max level (level 5).}
    \label{fig:vs_octree}
\end{figure*}

\begin{figure*}[ht]
    \includegraphics[width=\textwidth]{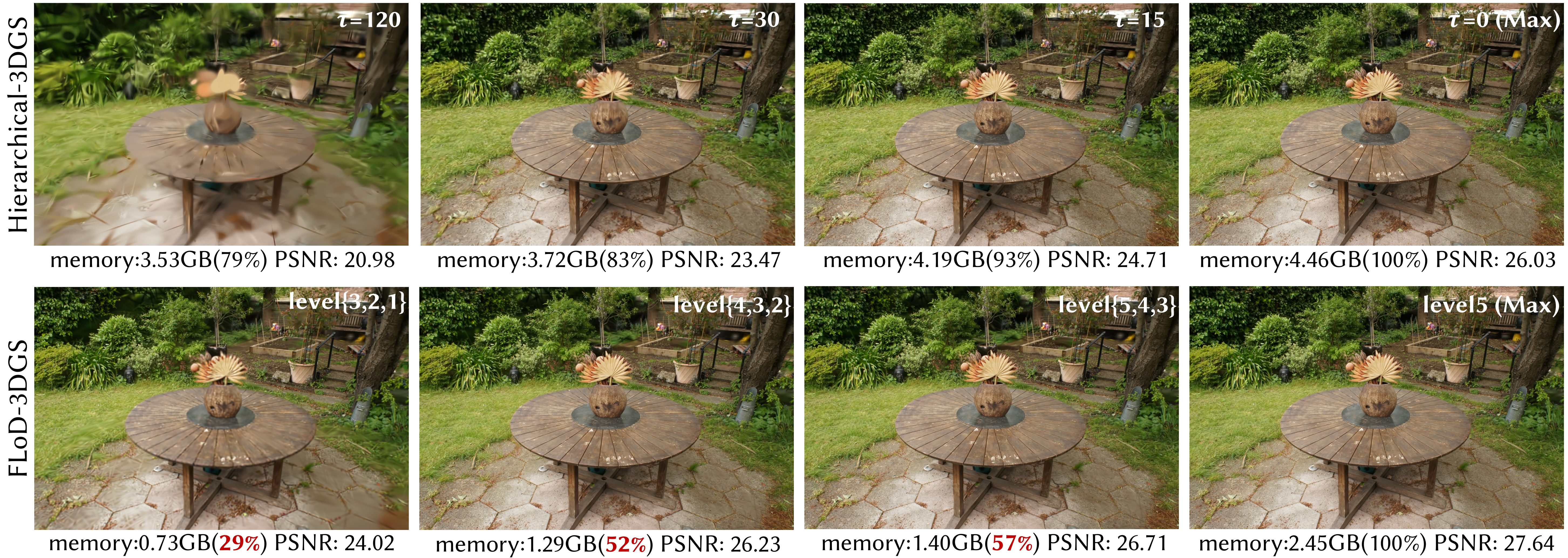}
    \caption{
    Comparison of the trade-off between visual quality and memory usage for FLoD-3DGS and Hierarchical-3DGS. The percentages (\%) shown next to the memory values indicate how much memory each rendering setting consumes relative to the memory required by the "Max" setting for maximum rendering quality.
    }
    \label{fig:selective_garden}
\end{figure*}

\subsubsection{Implementation}

FLoD-3DGS is implemented on the 3DGS framework. Experiments are mainly conducted on a single NVIDIA RTX A5000 24GB GPU.
Following the common practice for LoD in graphics applications, we train our FLoD representation up to level \(L_\text{max} = 5 \).
Note that \(L_\text{max}\) is adjustable for specific objectives and settings with minimal impact on render quality.
For FLoD-3DGS training with \(L_\text{max}=5\) levels, we set the training iterations for levels 1, 2, 3, 4, and 5 to 10,000, 15,000, 20,000, 25,000, and 30,000, respectively. 
The number of training iterations for the max level matches that of the backbone, while the lower levels have fewer iterations due to their faster convergence.

Gaussian density control techniques (densification, pruning, overlap pruning, opacity reset) are applied during the initial 5,000, 6,000, 8,000, 10,000, and 15,000 iterations for levels 1, 2, 3, 4, and 5, respectively. 
The Gaussian density control techniques run for the same duration as the backbone at the max level, but for shorter durations at the lower levels, as fewer Gaussians need to be optimized.
Additionally, the intervals for densification are set to 2,000, 1,000, 500, 500, and 200 iterations for levels 1, 2, 3, 4, and 5, respectively. 
We use longer intervals compared to the backbone, which sets the interval to 100, as to allow more time for Gaussians to be optimized before new Gaussians are added or existing Gaussians are removed. 
These settings were selected based on empirical observations.
Overlap pruning runs every 1000 iterations at all levels except the max level, where it is not applied.

We set the initial 3D scale constraint \(\lambda\) to 0.2 and the scale factor \(\rho\) to 4.  
This configuration effectively distinguishes the level of detail across \(L_\text{max}\) levels in most of the scenes we handle, enabling LoD representations that adapt to various memory capacities.  
For smaller scenes or when higher detail is required at lower levels, the initial 3D scale constraint \(\lambda\) can be further reduced.    

Unlike the original 3DGS approach, we do not periodically remove large Gaussians or those with large projected sizes during training as we do not impose an upper bound on the Gaussian scale. 
All other training settings not mentioned follow those of the backbone model. For loss, we adopt L1 and SSIM losses across all levels, consistent with the backbone model.

For selective rendering, we default to using the predetermined Gaussian set unless stated otherwise. 
The screen size threshold \(\gamma\) is set as 1.0. 
This selects Gaussians of level \(l\) from distances where the image projection of the level-specific 3D scale constraint \(s_\text{min}^{(l)}\) becomes equal or smaller than 1.0 pixel length. 



\subsection{Flexible Rendering}

In this section, we show that each level representation from FLoD can be used independently.
Based on this, we demonstrate the extensive range of rendering options that FLoD offers, through both single and selective rendering. 

\begin{figure*}[t]
    \includegraphics[width=\textwidth]{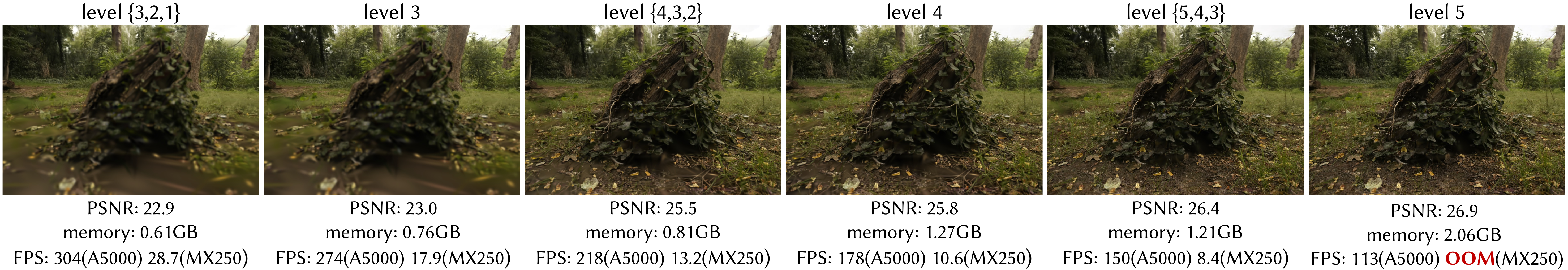}
    \caption{
    Various rendering options of FLoD-3DGS are evaluated on a server with an A5000 GPU and a laptop equipped with a 2GB VRAM MX250 GPU. 
    The flexibility of FLoD-3DGS provides rendering options that prevent out-of-memory (OOM) errors and allow near real-time rendering on the laptop setting.
    }
    \label{fig:various_stump}
\end{figure*}

\subsubsection{LoD Representation}
\label{sec:lod_represent}

As shown in Figure~\ref{fig:single}, 
FLoD follows the LoD concept by offering independent representations at each level. Each level captures the scene with varying levels of detail and corresponding memory requirements.
This enables users to select an appropriate level for rendering based on the desired visual quality and available memory.
A key observation is that even at lower levels (e.g., levels 1, 2, and 3), FLoD-3DGS achieves high perceptual visual quality for the background. 
This is because, even with the large size of Gaussians at lower levels, the perceived detail in distant regions is similar to that achieved using the smaller Gaussians at higher levels.

To further demonstrate the effectiveness of FLoD’s level representations, we compare renderings of each level from FLoD-3DGS with those from Octree-3DGS, as shown in Figure~\ref{fig:vs_octree}. 
At lower levels (e.g., levels 1, 2, and 3), Octree-3DGS shows broken structures, such as a pavilion, and the sharp artifacts created by very thin and elongated Gaussians.
In contrast, FLoD-3DGS preserves the overall structure with appropriate detail for each level. 
Notably, it achieves this while using fewer Gaussians than Octree-3DGS, showing our method's superiority in efficiently creating lower-level representations that better capture the scene structure. 
At higher levels (e.g., level 5), FLoD-3DGS uses more Gaussians to achieve higher visual quality and accurately reconstruct complex scene structures. 
This shows that our method can handle detailed scenes effectively through the higher level representations. 

In summary, the level representations of FLoD-3DGS outperform those of Octree-3DGS in reconstructing scene structures, as evidenced by its higher SSIM values across all levels.
Furthermore, FLoD-3DGS uses significantly fewer Gaussians at lower levels, requiring only 0.7\%, 2\%, and 22\% of the Gaussians of the max level for levels 1, 2, and 3, respectively. 
These results demonstrate that FLoD-3DGS can create level representations with a wide range of memory requirements.

Note that we exclude Hierarchical-3DGS from this comparison because it was not designed for rendering with specific levels.
For render results of Hierarchical-3DGS and Octree-3DGS that use Gaussians from single levels individually, please refer to Appendix~\ref{sec:appendix_single}.

\begin{figure}[t]
    \includegraphics[width=\columnwidth]{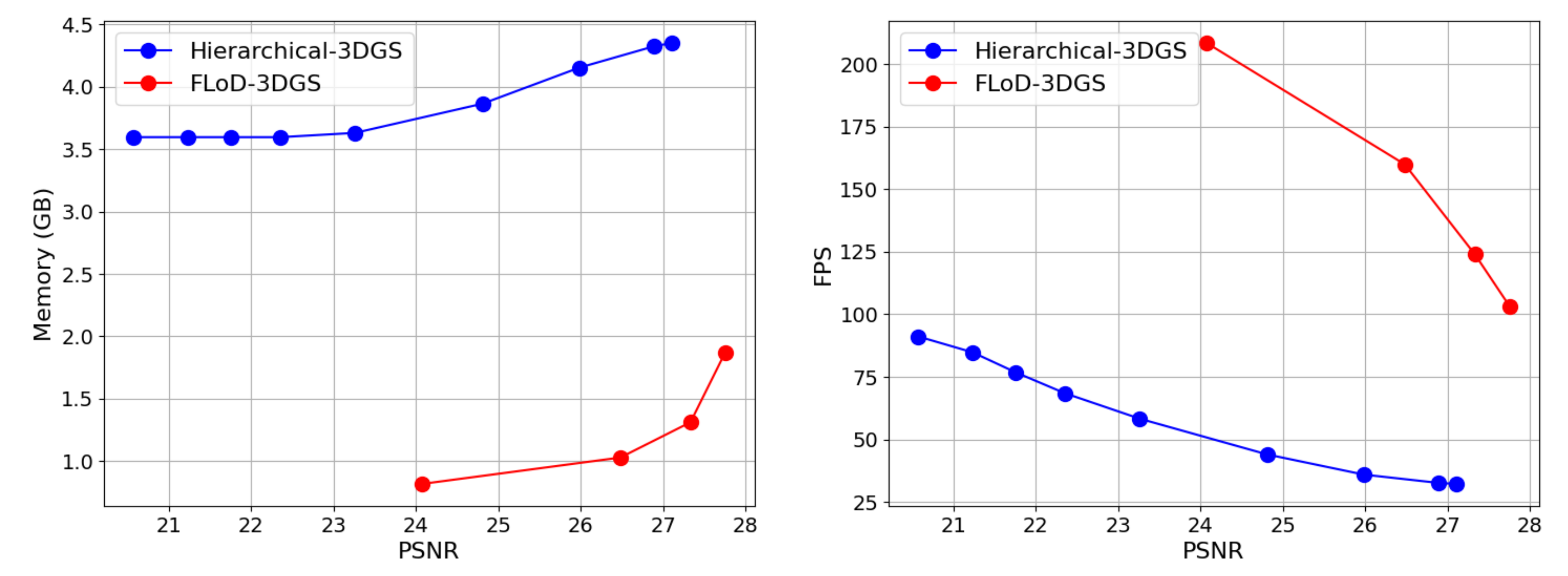}
    \caption{
    Comparison of the trade-offs in selective rendering for FLoD-3DGS and Hierarchical-3DGS on Mip-NeRF360 scenes: visual quality(PSNR) versus memory usage, and visual quality versus rendering speed(FPS). 
    }
    \label{fig:compare_graphs}
\end{figure}

\subsubsection{Selective Rendering}
\label{sec:selective_rendering_exp}
FLoD provides not only single-level rendering but also selective rendering. 
Selective rendering enables more efficient rendering by selectively using Gaussians from multiple levels.

To evaluate the efficiency of FLoD's selective rendering, we compare rendering quality and memory usage for different selective rendering configurations against Hierarchical-3DGS. 
We compare with Hierarchical-3DGS because its rendering method, involving the selection of Gaussians from its hierarchy based on target granularity \(\tau\), is similar to our selective rendering which selects Gaussians across level ranges based on the screen size threshold \(\gamma\).

As shown in Figure~\ref{fig:selective_garden}, FLoD-3DGS effectively reduces memory usage through selective rendering. 
For example, selectively using levels 5, 4, and 3 reduces memory usage by about half compared to using only level 5, while the PSNR decreases by less than 1. Similarly, selective rendering with levels 3, 2, and 1 reduce memory usage to approximately 30\%, with PSNR drop of about 3.6.

In contrast, Hierarchical-3DGS does not reduce memory usage as effectively as FLoD-3DGS and also suffers from a greater decrease in rendering quality.
Even when the target granularity \(\tau\) is set to 120, occupied GPU memory remains high, consuming approximately 79\% of the memory used for the maximum rendering quality setting (\(\tau=0\)). 
Moreover, for this rendering setting, the PSNR drops significantly by more than 5. 
These results demonstrate that FLoD-3DGS's selective rendering provides a wider range of rendering options, achieving a better balance between visual quality and memory usage compared to Hierarchical-3DGS.

We further compare the memory usage to PSNR curve, and FPS to PSNR curve on the Mip-NeRF360 scenes in Figure~\ref{fig:compare_graphs}.
For FLoD-3DGS, we evaluate rendering performance using only level 5, as well as selectively using levels 5, 4, 3; levels 4, 3, 2; and levels 3, 2, 1. 
For Hierarchical-3DGS, we measure rendering performance with target granularity \(\tau\) set to 0, 6, 15, 30, 60, 90, 120, 160, and 200.
The results show that FLoD-3DGS consistently uses less memory and achieves higher fps than Hierarchical-3DGS when compared at the same PSNR levels. 
Notably, as PSNR decreases, FLoD-3DGS shows a sharper reduction in memory usage, and a greater increase in fps.

Note that for a fair comparison, we train Hierarchical-3DGS with a maximum \(\tau\) of 200 during the hierarchy optimization stage to enhance its rendering quality for larger \(\tau\) beyond its default settings. 
For renderings of Hierarchicial-3DGS using its default training settings, please refer to Appendix~\ref{sec:appendix_selective}.

\input{tables/quant_baselines}
\input{tables/selective_mipnerf}

\subsubsection{Various Rendering Options}

FLoD supports both single-level rendering and selective rendering, offering a wide range of rendering options with varying visual quality and memory requirements. 
As shown in Table~\ref{tab:selective_mipnerf}, FLoD enables flexible adjustment of the number of Gaussians. Reducing the number of Gaussians increases rendering speed while also reducing memory usage, allowing FLoD to adapt efficiently to hardware environments with varying memory constraints.

To evaluate the flexibility of FLoD, we conduct experiments on a server with an A5000 GPU and a low-cost laptop equipped with a 2GB VRAM MX250 GPU. As shown in Figure~\ref{fig:various_stump}, rendering with only level 4 or selective rendering using levels 5, 4, and 3 achieves visual quality comparable to rendering with only level 5, while reducing memory usage by approximately 40\%. This reduction prevents out-of-memory (OOM) errors that occur on low-cost GPUs, such as the MX250, when rendering with only level 5. Furthermore, using lower levels for single-level rendering or selective rendering increases fps, enabling near real-time rendering even on low-cost devices.

Hence, FLoD offers considerable flexibility by providing various rendering options through single and selective rendering, ensuring effective performance across devices with different memory capacities. 
For additional evaluations of rendering flexibility on the MX250 GPU in Mip-NeRF360 scenes, please refer to the Appendix~\ref{sec:appendix_lowcost}.

\subsection{Max Level Rendering}
\label{sec:max_level}

We have demonstrated that FLoD provides various rendering options following the LoD concept. 
However, in this section, we show that using only the max level for single-level rendering provides rendering quality comparable to those of existing models.
Moreover, FLoD provides rendering quality comparable to those of existing models when using the maximum level for single-level rendering.
Table~\ref{tab:quant_table} compares FLoD-3DGS with baselines across three real-world datasets. Table~\ref{tab:quant_table} compares max-level (level 5) of FLoD-3DGS with baselines across three real-world datasets.

FLoD-3DGS performs competitively on the Mip-NeRF360 and Tanks\&Temples datasets, which are commonly used in baseline evaluations, and outperforms all baselines across all reconstruction metrics on the DL3DV-10K dataset. 
This demonstrates that FLoD achieves high-quality rendering, which users can select from among the various rendering options FLoD provides.
For qualitative comparisons, please refer to Appendix~\ref{sec:appendix_maxlevel_rendering}.

\begin{figure}[t]
    \includegraphics[width=\columnwidth]{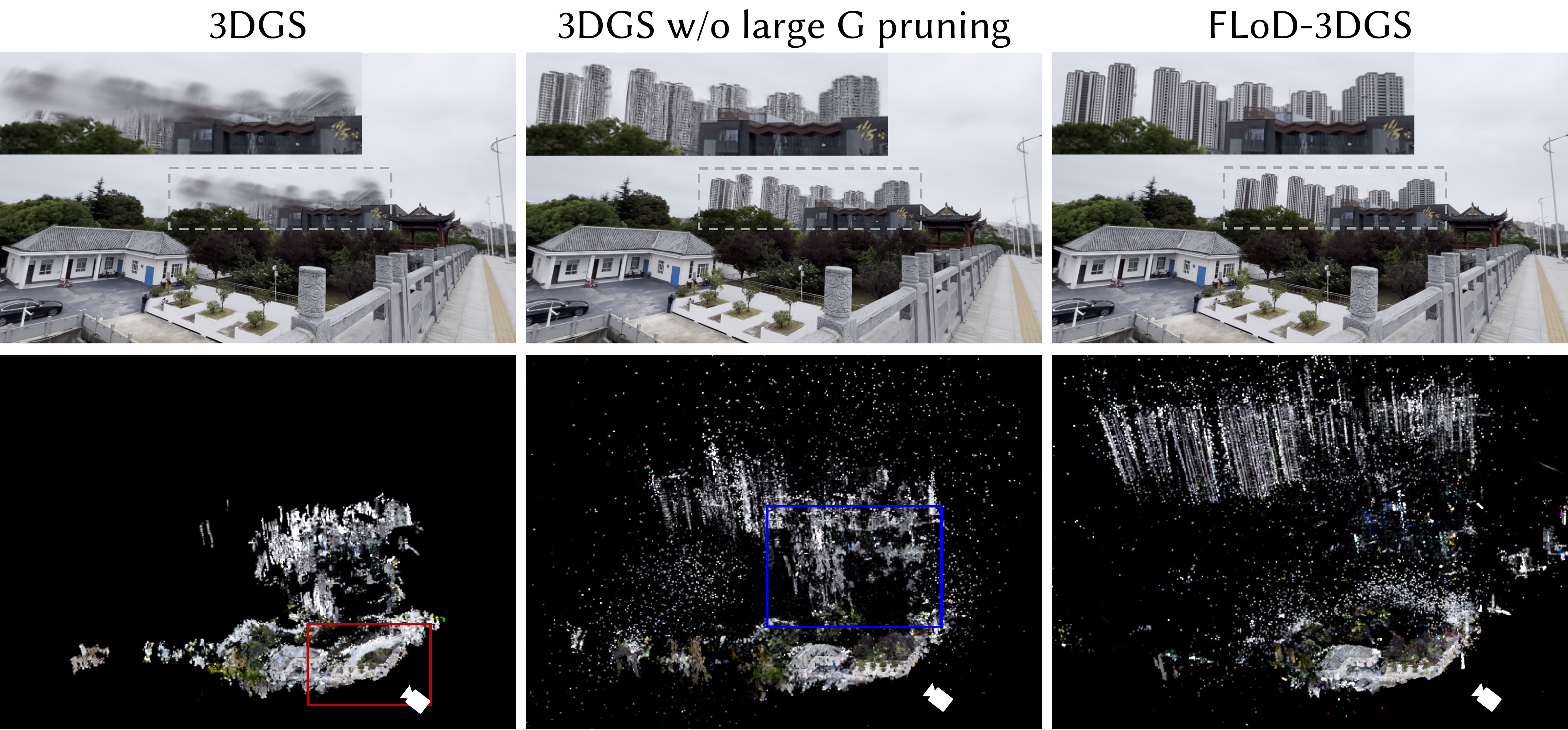}
    \caption{
    Comparison of 3DGS and FLoD-3DGS on the DL3DV-10K dataset. The upper row shows rendering with zoom-in of the gray dashed box. The bottom row shows point visualization of the Gaussian centers. The red box shows distortions caused by large Gaussian pruning, and the blue box illustrates geometry inaccuracies that occur without the 3D scale constraint. FLoD's 3D scale constraint ensures accurate Gaussian placement and improved rendering.
    }
    \label{fig:max_scsize}
\end{figure}

\paragraph{Discussion on rendering quality improvement}
FLoD-3DGS particularly excels at rendering high-quality distant regions. This results in high PSNR on the DL3DV-10K dataset, which contains many distant objects.
Two key differences from vanilla 3DGS drive this improvement: removing large Gaussian pruning and introducing a 3D scale constraint.

Vanilla 3DGS prunes large Gaussians during training. This pruning causes distant backgrounds, such as the sky and buildings, to be incorrectly rendered with small Gaussians near the camera, as shown in the red box in Figure~\ref{fig:max_scsize}.  This distortion disrupts the structure of the scene. Simply removing this pruning alleviates the problem and improves the rendering quality.

However, removing large Gaussian pruning alone does not guarantee accurate Gaussian placement. As shown in the blue box in Figure~\ref{fig:max_scsize}, buildings are rendered with Gaussians of varying sizes at different depths, resulting in inaccurate geometry in the rendered image.

FLoD's 3D scale constraint solves this issue. It initially constrains Gaussians to be large, applying greater loss to mispositioned Gaussians to correct or prune them. During training, densification adds new Gaussians near existing ones, preserving accurate geometry as training progresses. This approach allows FLoD to reconstruct scene structures more precisely and in the correct positions.

\subsection{Backbone Compatibility}

\input{tables/scaffold_comp}

Our method, FLoD, integrates seamlessly with 3DGS and its variants.
To demonstrate this, we apply FLoD not only to 3DGS (FLoD-3DGS) but also to Scaffold-GS that uses anchor-based neural Gaussians (FLoD-Scaffold).
As shown in Figure~\ref{fig:single}, FLoD-Scaffold also generates representations with appropriate levels of detail and memory for each level.

To further illustrate how FLoD-Scaffold provides suitable representations for each level across different datasets, we measure the PSNR and rendering memory usage for each level on three datasets. As shown in Table~\ref{tab:scaffold_comp}, FLoD-Scaffold provides various rendering options that balance visual quality and memory usage across all three datasets. In contrast, Octree-Scaffold, which also uses Scaffold-GS as its backbone model, has limitations in providing multiple rendering options due to its restricted representation capabilities for middle and low levels, similar to Octree-3DGS.

Furthermore, FLoD-Scaffold also shows high visual quality when rendering with only the max level (level 5). 
As shown in Table~\ref{tab:scaffold_comp}, FLoD-Scaffold outperforms Scaffold-GS and achieves competitive results with Octree-Scaffold across all datasets.

Consequently, FLoD can seamlessly integrate into existing 3DGS-based models, providing LoD functionality without degrading rendering quality. 
Furthermore, we expect FLoD to be compatible with future 3DGS-based models as well.

\subsection{Urban Scene}
\label{sec:urban_scene}

\begin{figure}[t]
    \includegraphics[width=\columnwidth]{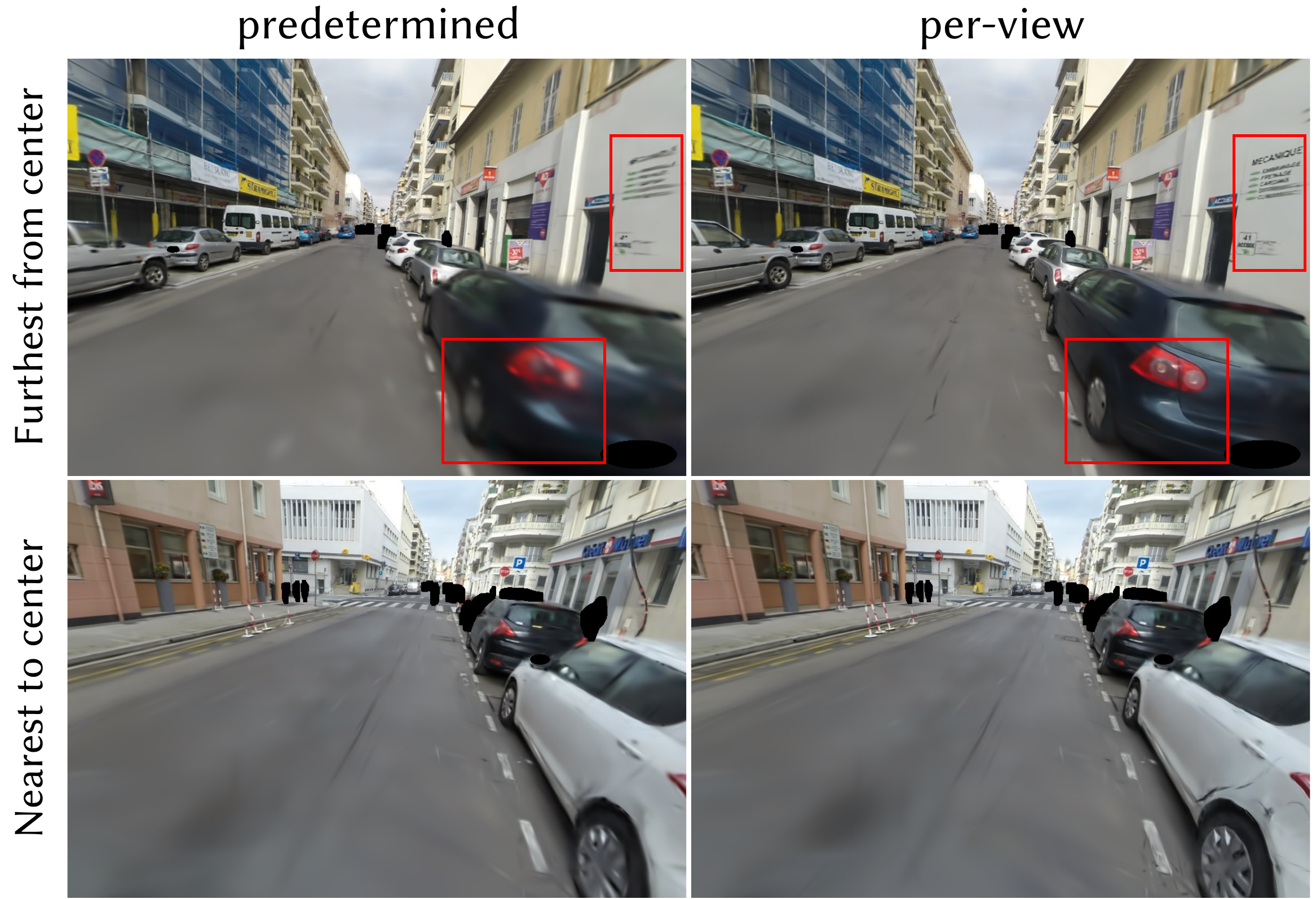}
    \caption{
    Comparison between the predetermined method and the per-view method in selective rendering using levels 5, 4, and 3 on the Small City scene. As shown in the red boxed areas, the per-view method maintains superior rendering quality even when far from the center of the scene, whereas the predetermined method shows a decline in rendering quality.
    }
    \label{fig:urbanscene}
\end{figure}

We further evaluate our method on Small City scene~\cite{kerbl2024hierarchicalgaussians}, which is a scene collected in Hierachcial-3DGS for evaluation.
In urban scenes, where cameras cover extensive areas, selective rendering with a predetermined Gaussian set \(\mathbf{G}_\text{sel}\) can result in noticeable decline in rendering detail. 
This problem arises because the predetermined Gaussian set allocates higher level Gaussians around the average training camera position and lower levels for more distant areas. 
Consequently, as the camera moves into these peripheral areas, the rendering quality drops as lower level Gaussians are rasterized near the camera. 
Figure~\ref{fig:urbanscene}(left) shows that predetermined Gaussian set \(\mathbf{G}_\text{sel}\) cannot maintain rendering quality when the camera moves far from this central position.

\input{tables/urban_scene}

To maintain rendering quality across varying camera positions in urban environments, it is necessary to dynamically adapt the Gaussian set \(\mathbf{G}_\text{sel}\).
As shown in Figure~\ref{fig:urbanscene}(right), selective rendering with per-view Gaussian set \(\mathbf{G}_\text{sel}\) maintains consistent rendering quality. 
Compared to using the predetermined \(\mathbf{G}_\text{sel}\), per-view \(\mathbf{G}_\text{sel}\) increases PSNR by 0.8, but with a slower rendering speed and more rendering memory demands (Table~\ref{tab:urbanscene}). 
The slowdown occurs because the rendering of each view has an additional process of creating \(\mathbf{G}_\text{sel}\).
To mitigate the reduction in rendering speed, all Gaussians within the level range [\(L_\text{start}\), \(L_\text{end}\)] are kept in GPU memory, which accounts for the increased memory usage.
Despite the drawbacks, the trade-off for per-view \(\mathbf{G}_\text{sel}\) selective rendering is considered reasonable as the rendering quality becomes consistent, and it offers a faster rendering option compared to max level rendering.

Table~\ref{tab:urbanscene} also shows that our selective rendering (per-view) method not only achieves better PSNR with a comparable number of Gaussians but also outperforms Hierarchical-3DGS (\(\tau=30\)) in efficiency. 
Although both methods create the Gaussians set \(\mathbf{G}_\text{sel}\) for every individual view, our method achieves faster FPS and uses less rendering memory. 

\subsection{Ablation Study}

\subsubsection{3D Scale Constraint}

\begin{figure}[t]
    \includegraphics[width=\columnwidth]{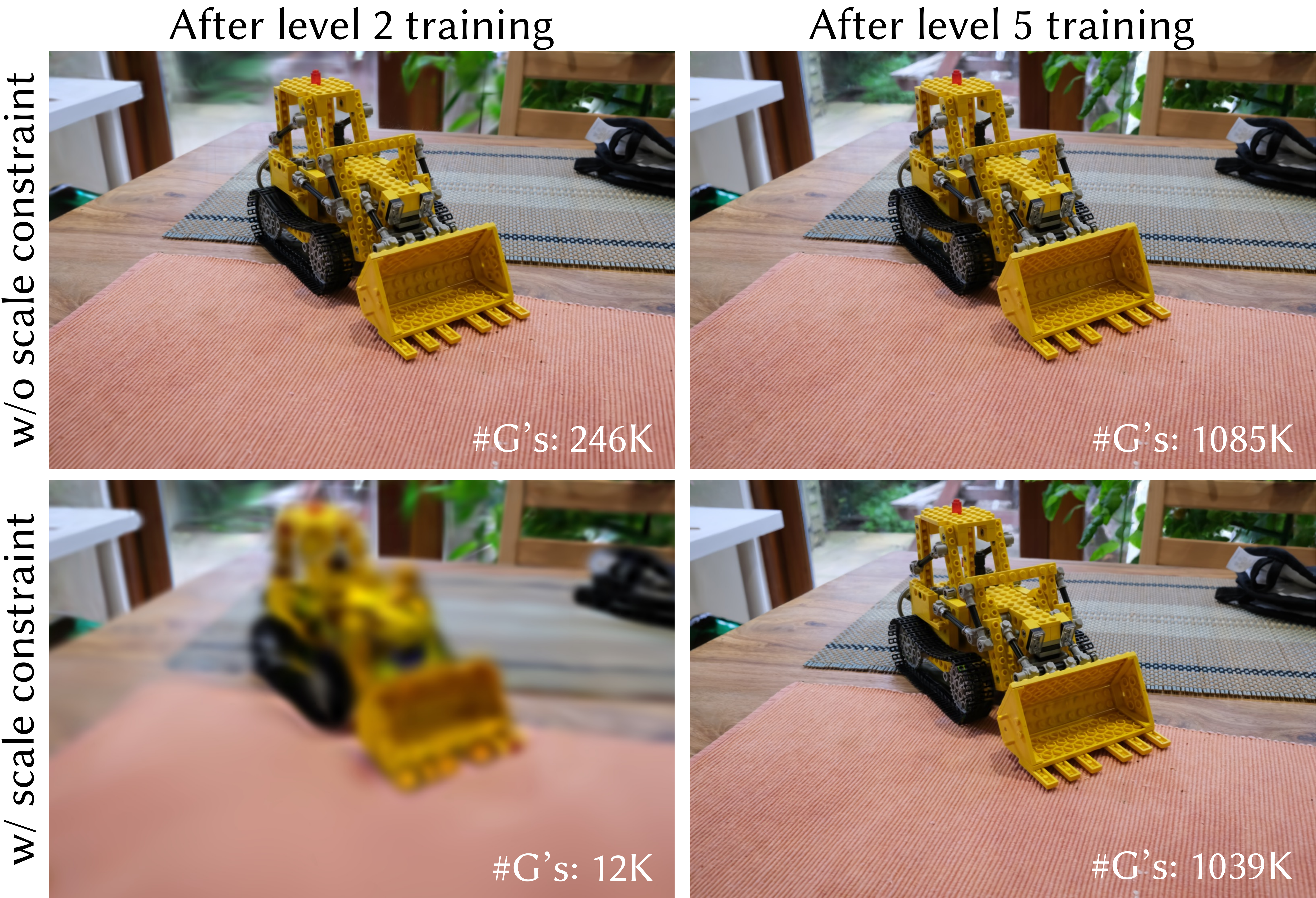}
    \caption{
    Comparison of the renderings and number of Gaussians with and without the 3D scale constraint after level 2 and level 5 training on the Mip-NeRF360 dataset.
    }
    \label{fig:size_constraint}
\end{figure}

We compare cases with and without the 3D scale constraint. 
For the case without the 3D scale constraint, Gaussians are optimized without any size limit. 
Additionally, we did not apply overlap pruning for this case, as the threshold for overlap pruning \( d_\text{OP}^{(l)} \) is adjusted proportionally to the 3D scale constraint. 
Therefore, the case without the 3D scale constraint only applies level-by-level training method from our full method.

As shown in Figure~\ref{fig:size_constraint}, without the 3D scale constraint, the amount of detail reconstructed after level 2 is comparable to that after the max level. 
In contrast, applying the 3D scale constraint results in a clear difference in detail between the two levels. 
Moreover, the case with the 3D scale constraint uses approximately 98.6\% fewer Gaussians compared to the case without the 3D scale constraint.
Therefore, the 3D scale constraint is crucial for ensuring varied detail across levels and enabling each level to maintain a different memory footprint.

\begin{figure}[t]
    \includegraphics[width=\columnwidth]{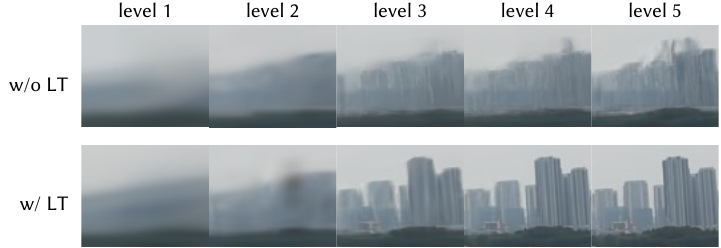}
    \caption{
    Comparison of background region on the rendered images with and without level-by-level training across all levels on the DL3DV-10K dataset. 
    The images are zoomed-in and cropped to highlight differences in the background regions.
    }
    \label{fig:level_train}
\end{figure}

\subsubsection{Level-by-level Training}
\input{tables/abl_lbl}
\begin{figure}[t]
    \includegraphics[width=\columnwidth]{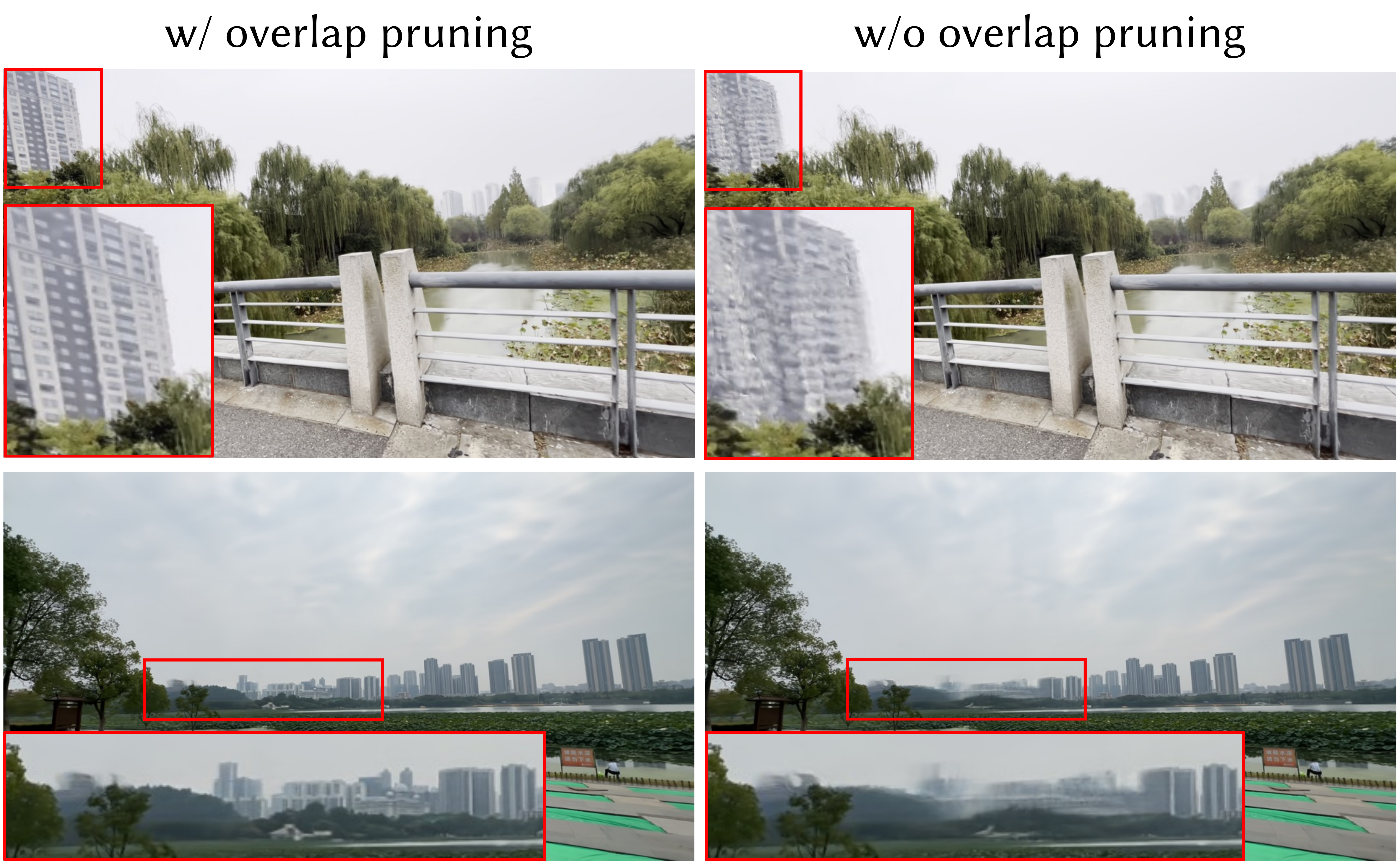}
    \caption{
    Comparison between rendered images at level 5 trained with and without overlap pruning on the DL3DV-10K dataset. Zoomed-in images emphasize key differences.
    }
    \label{fig:overlap}
\end{figure}

We compare cases with and without the level-by-level training approach. In the case without level-by-level training, the set of iterations for exclusive Gaussian optimization of each level is replaced with iterations that include additional densification and pruning. As shown in Figure~\ref{fig:level_train}, the absence of level-by-level training causes inaccuracies in the reconstructed structure at the intermediate level, which is carried on to the higher levels.

In contrast, the case with our level-by-level training approach reconstructs the scene structure more accurately at level 3, resulting in improved reconstruction quality at levels 4 and 5. As demonstrated in Table~\ref{tab:lbl_all}, the case with level-by-level training outperforms the case without it in terms of PSNR, SSIM, and LPIPS across all levels.
Hence, level-by level training is important for enhancing reconstruction quality across all levels.

\subsubsection{Overlap Pruning}

We compare the result of training with and without overlap pruning across all levels.
As shown in Figure~\ref{fig:overlap}, removing overlap pruning deteriorates the structure of the scene, degrading rendering quality.
This issue is particularly noticeable in scenes with distant objects. 
We believe that overlap pruning mitigates the potential for artifacts by preventing the overlap of large Gaussians at distant locations.

Furthermore, we compare the number of Gaussians at each level with and without overlap pruning. 
Table~\ref{tab:overlap} illustrates that overlap pruning decreases the number of Gaussians, particularly at lower levels, with reductions of 90\%, 34\%, and 10\% at levels 1, 2, and 3, respectively.
This reduction is particularly important for minimizing memory usage for rendering on low-cost and low-memory devices that utilize low level representations.

\input{tables/abl_overlap}

\section{Conclusion}
In this work, we propose Flexible Level of Detail (FLoD), a method that integrates LoD into 3DGS. 
FLoD reconstructs the scene in different degrees of detail while maintaining a consistent scene structure.
Therefore, our method enables customizable rendering with a single or subset of levels, allowing the model to operate on devices ranging from high-end servers to low-cost laptops.
Furthermore, FLoD easily integrates with 3DGS-based models implying its applicability to future 3DGS-based methods.

\section{Limitation}

In scenes with long camera trajectories, using per-view Gaussian set is necessary to maintain consistent rendering quality during selective rendering. 
However, this method has the limitation that all Gaussians within the level range for selective rendering need to be kept on GPU memory to maintain fast rendering rates, as discussed in Section~\ref{sec:urban_scene}.
Therefore, this method requires more memory capacity compared to single level rendering with only the highest level, \(L_\text{end}\), picked from the level range [\(L_\text{start}\), \(L_\text{end}\)] used for selective rendering. 
Future research could explore the strategic planning and execution of transferring Gaussians from the CPU to the GPU, 
to reduce the memory burden while also keeping the advantage of selective rendering.

\begin{acks}
This work was supported by the National Research Foundation of Korea (NRF, RS-2023-00223062) and an IITP grant (RS-2020-II201361, Artificial Intelligence Graduate School Program (Yonsei University)) funded by the Korean government (MSIT) .
\end{acks}

%% file: tables/quant_baselines.tex

\begin{table*}[ht]
\caption{Quantitative comparison of FLoD-3DGS to baselines across three real-world datasets (Mip-NeRF360, DL3DV-10K, Tanks\&Temples). For FLoD-3DGS and Hierarchical-3DGS, we use the rendering setting that produces the best image quality. The best results are highlighted in bold.
}
\centering
\begin{tabular}{l|ccc|ccc|ccc}
\hline
\multicolumn{1}{c|}{} & \multicolumn{3}{c|}{Mip-NeRF360} & \multicolumn{3}{c|}{DL3DV-10K} & \multicolumn{3}{c}{Tanks\&Temples} \\

\text{} & $\text{PSNR}\uparrow$ & $\text{SSIM}\uparrow$ & $\text{LPIPS}\downarrow$ & $\text{PSNR}\uparrow$ & $\text{SSIM}\uparrow$ & $\text{LPIPS}\downarrow$ & $\text{PSNR}\uparrow$ & $\text{SSIM}\uparrow$ & $\text{LPIPS}\downarrow$ \\
\hline
3DGS & 27.36 & 0.812 & 0.217 & 28.00 & 0.908 & 0.142 & 23.58 & 0.848 & 0.177 \\

Mip-Splatting & 27.59 & \textbf{0.831} & \textbf{0.181} & 28.64 & 0.917 & 0.125 & 23.62 & 0.855 & 0.157 \\

Octree-3DGS & 27.29 & 0.815 & 0.214 & 29.14 & 0.915 & 0.128 & 24.19 & \textbf{0.865} & 0.154 \\

Hierarchical-3DGS & 27.10 & 0.797 & 0.219 & 30.45 & 0.922 & 0.115 & 24.03 & 0.861 & \textbf{0.152} \\

FLoD-3DGS & \textbf{27.75} & 0.815 & 0.224 & \textbf{31.99} & \textbf{0.937} & \textbf{0.107} & \textbf{24.41} & 0.850 & 0.186  \\
\hline
\end{tabular}

\label{tab:quant_table}
\end{table*}


%% file: tables/selective_mipnerf.tex

\begin{table}[t]
\caption{
Trade-offs between visual quality, rendering speed, and the number of Gaussians achieved in FLoD-3DGS through single-level and selective rendering in the Mip-NeRF360 dataset.
}
\centering
\begin{tabular}{ccccc|ccc|c|c}
\hline
\multicolumn{5}{c|}{Level} & & & & & \\ 
1 & 2 & 3 & 4 & 5 & $\text{PSNR}$ & $\text{SSIM}$ & $\text{LPIPS}$ & $\text{FPS}$ & $\text{\#G's}$ \\
\hline
\multicolumn{5}{r|}{\cmark} & 27.75 & 0.815 & 0.224 & 103 & 2189K \\
\multicolumn{5}{r|}{\cmark - \cmark - \cmark} & 27.33 & 0.801 & 0.245 & 124 & 1210K \\
\multicolumn{5}{r|}{\cmark \phantom{$-\checkmark$}} & 26.67 & 0.764 & 0.292 & 150 & 1049K \\
\multicolumn{5}{c|}{\cmark - \cmark - \cmark}& 26.48 & 0.759 & 0.298 & 160 & 856K \\
\multicolumn{5}{c|}{\cmark} & 24.11 & 0.634 & 0.440 & 202 & 443K \\
\multicolumn{5}{l|}{\cmark - \cmark - \cmark}& 24.07 & 0.632 & 0.442 & 208 & 414K \\
\hline
\end{tabular}

\label{tab:selective_mipnerf}
\end{table}


%% file: tables/scaffold_comp.tex
\begin{table}[t]
\caption{
Level-wise comparison of visual quality and memory usage (GB) for FLoD-3DGS, alongside Scaffold-GS and Octree-GS on Mip-NeRF360(Mip), DL3DV-10K(DL3DV) and Tanks\&Temples(T\&T) datasets.
}
\centering
\setlength{\tabcolsep}{1mm}
\begin{tabular}{l|cc|cc|cc}
\hline
\multicolumn{1}{c|}{} & \multicolumn{2}{c|}{Mip} & \multicolumn{2}{c|}{DL3DV} & \multicolumn{2}{c}{T\&T} \\
& $\text{PSNR}$ & mem. & $\text{PSNR}$ & mem. & $\text{PSNR}$ & mem. \\
\hline
FLoD-Scaffold(lv1) & 20.1 & 0.5 & 22.2 & 0.3 & 17.1 & 0.2 \\
FLoD-Scaffold(lv2) & 22.1 & 0.5 & 25.2 & 0.3 & 19.3 & 0.3 \\
FLoD-Scaffold(lv3) & 24.7 & 0.6 & 28.5 & 0.4 & 21.8 & 0.4 \\
FLoD-Scaffold(lv4) & 26.6 & 0.8 & 30.1 & 0.6 & 23.6 & 0.7 \\
FLoD-Scaffold(lv5) & 27.4 & 1.0 & 31.1 & 0.7 & 24.1 & 1.0 \\
\hline
Scaffold-GS & 27.4 & 1.3 & 30.5 & 0.8 & 24.1 & 0.7 \\
Octree-Scaffold & 27.2 & 1.0 & 30.9 & 0.6 & 24.6 & 0.8 \\
\hline
\end{tabular}

\label{tab:scaffold_comp}
\end{table}

%% file: tables/urban_scene.tex
\begin{table}[t]
\caption{
Quantitative comparison of FLoD-3DGS to Hierarchical-3DGS in Small City scene. 
The upper section compares FLoD-3DGS's selective rendering methods and Hierarchical-3DGS (\(\tau=30\)), where all methods use a similar number of Gaussians. Note that \#G's for our per-view method and Hierarchical-3DGS is based on the view using the most number of Gaussians as this number varies across different views. The lower section lists the maximum quality renderings for both FLoD-3DGS and Hierarchical-3DGS for comparison.
}
\centering
\begin{tabular}{l|ccc|c}
\hline
 & $\text{PSNR}\uparrow$ & $\text{FPS}\uparrow$ & mem.$\downarrow$ & $\text{\#G's}\downarrow$ \\
\hline
FLoD-3DGS \small{(per-view)} & \textbf{25.49} & 221 & 1.03\small{GB} & 601K \\
FLoD-3DGS \small{(predetermined)} & 24.69 & \textbf{286} & \textbf{0.41\small{GB}} & \textbf{589K} \\
Hierarchcial-3DGS \small{($\tau=30$)} & 24.69 & 55 & 5.36\small{GB} & 610K \\
\hline
FLoD-3DGS \small{(max level)} & 26.37 & 181 & 0.86\small{GB} & 1308K \\
Hierarchcial-3DGS \small{($\tau=0$)} & 26.69 & 17 & 7.81\small{GB} & 4892K \\
\hline
\end{tabular}
\label{tab:urbanscene}
\end{table}

%% file: tables/abl_lbl.tex
\begin{table}[t]
\caption{
Quantitative comparison of image quality for each level with and without level-by-level training on DL3DV-10K dataset. LT denotes level-by-level training.
}
\centering
\begin{tabular}{c|l|ccc}
\hline
Level & Training & $\text{PSNR}\uparrow$ & $\text{SSIM}\uparrow$ & $\text{LPIPS}\downarrow$ \\
\hline
5 & w/o LT & 31.20 & 0.930 & 0.158 \\
& w/ LT & 31.97 & 0.936 & 0.105 \\
\hline
4 & w/o LT & 29.05 & 0.896 & 0.161 \\
& w/ LT & 30.73 & 0.917 & 0.133 \\
\hline
3 & w/o LT & 27.05 & 0.850 & 0.224 \\
& w/ LT & 28.29 & 0.869 & 0.200 \\
\hline
2 & w/o LT & 23.41 & 0.734 & 0.376 \\
& w/ LT  & 24.01 & 0.750 & 0.355 \\
\hline
1 & w/o LT & 20.41 & 0.637 & 0.485 \\
& w/ LT  & 20.81 & 0.646 & 0.475 \\
\hline
\end{tabular}

\label{tab:lbl_all}
\end{table}

%% file: tables/abl_overlap.tex
\begin{table}[t]
\caption{Comparison of the number of Gaussians per level when trained with and without overlap pruning on the Mip-NeRF360 dataset. OP denotes overlap pruning.}
\centering

\begin{tabular}{l|ccccc}
\hline
Level & 1 & 2 & 3 & 4 & 5 \\
\hline
w/o OP & 38K & 49K & 439K & 1001K & 2058K \\
w/ OP & 10K & 31K & 390K & 970K & 2048K \\
\hline
\end{tabular}

\label{tab:overlap}
\end{table}


%% file: supple.tex
\appendix

\section{Dataset Details}
\label{sec:appendix_dataset}

We conduct experiments on the Tanks\&Temples dataset ~\cite{Knapitsch2017tandt} and the Mip-NeRF360 dataset~\cite{barron2022mipnerf360} as the two datasets were used for evaluation in our baselines: Octree-GS~\cite{ren2024octreegs}, 3DGS~\cite{kerbl3Dgaussians}, Scaffold-GS~\cite{lu2023scaffoldgs} and Mip-Splatting~\cite{yu2024mipsplatting}.
Additionally, we conduct experiments on the relatively recently released DL3DV-10K dataset~\cite{ling2023dl3dv10k} for a more comprehensive evaluation across diverse scenes.
Camera parameters and initial points for all datasets are obtained using COLMAP~\cite{schoenberger2016sfm}. 
We subsample every 8th image of each scene for testing, following the train/test splitting methodology presented in Mip-NeRF360.

\subsection{Tanks\&Temples}
The Tanks\&Temples dataset includes high-resolution multi-view images of various complex scenes, including both indoor and outdoor settings.
Following our baselines, we conduct experiments on two unbounded scenes featuring large central objects: train and truck. For both scenes, we reduce the image resolution to \(980 \times 545\) pixels, downscaling it to 25\% of their original size.

\subsection{Mip-NeRF360}
The Mip-NeRF360 dataset~\cite{barron2022mipnerf360} consists of a diverse set of real-world 360-degree scenes, encompassing both bounded and unbounded environments.
The images in the dataset were captured under controlled conditions to minimize lighting variations and avoid transient objects. 
For our experiments, we use the nine publicly available scenes: bicycle, bonsai, counter, garden, kitchen, room, stump, treehill and flowers. 
We reduce the original image's width and height to one-fourth for the outdoor scenes, and to one-half for the indoor scenes.
Specifically, the outdoor scenes are resized to approximately \(1250 \times 830\) pixels, while the indoor scenes are resized to about \(1558 \times 1039\) pixels.

\subsection{DL3DV-10K}
The DL3DV-10K dataset~\cite{ling2023dl3dv10k} expands the range of real-world scenes available for 3D representation learning by providing a vast number of indoor and outdoor real-world scenes.
For our experiments, we select six outdoor scenes from DL3DV-10K for a more comprehensive evaluation on unbounded real-world environments.
We use images with a reduced resolution of \(960 \times 540\) pixels, following the resolution used in the DL3DV-10K paper. 
The first 10 characters of the hash codes for our selected scenes are aeb33502d5, 58e78d9c82, df87dfc4c, ce06045bca, 2bfcf4b343, and 9f518d2669.

\begin{figure*}[ht]
    \includegraphics[width=\textwidth]{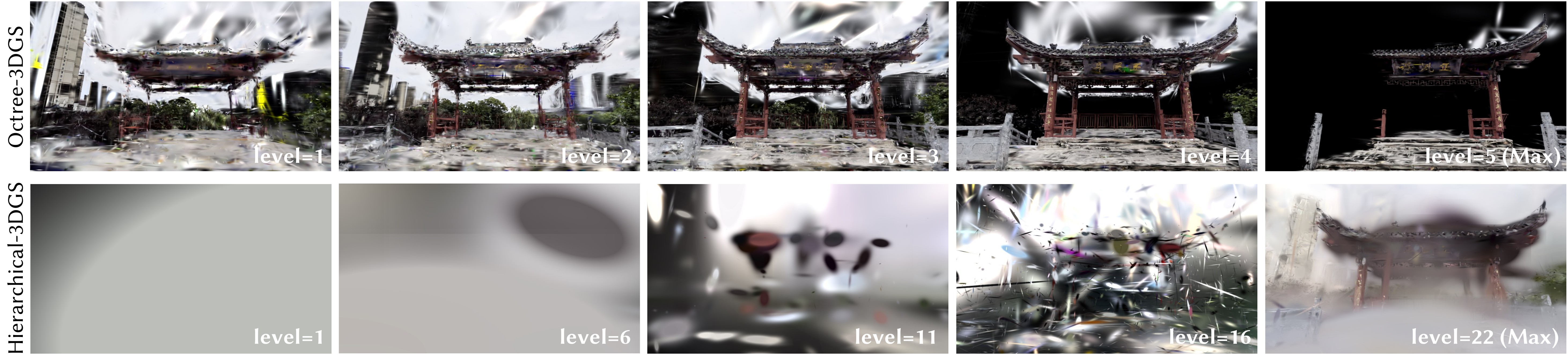}
    \caption{Rendered images using only the Gaussians corresponding to a specific level in Octree-3DGS and Hierarchical-3DGS.}
    \label{fig:appendix_per_level}
\end{figure*}

\input{tables/algorithm}

\section{Method Details}
\subsection{Training Algorithm}
The overall training process for FLoD-3DGS is summarized in Algorithm~\ref{tab:algorithm}.

\subsection{3D vs 2D Scale Constraint}
It is essential to impose the Gaussian scale constraint in 3D rather than on the 2D projected Gaussians. 
Although applying scale constraints to 2D projections is theoretically possible, it increases geometrical ambiguities in modeling 3D scenes. 
This is because the scale of the 2D projected Gaussians varies depending on their distance from the camera. 
Consequently, imposing a constant scale constraint on a 2D projected Gaussian from different camera positions sends inconsistent training signals, leading to Gaussian receiving training signals that misrepresent their true shape and position in 3D space.
In contrast, applying 3D scale constraint to 3D Gaussians ensures consistent enlargement regardless of the camera's position, thereby enabling stable optimization of the Gaussians' 3D scale and position.

\begin{figure*}[t]
    \includegraphics[width=\textwidth]{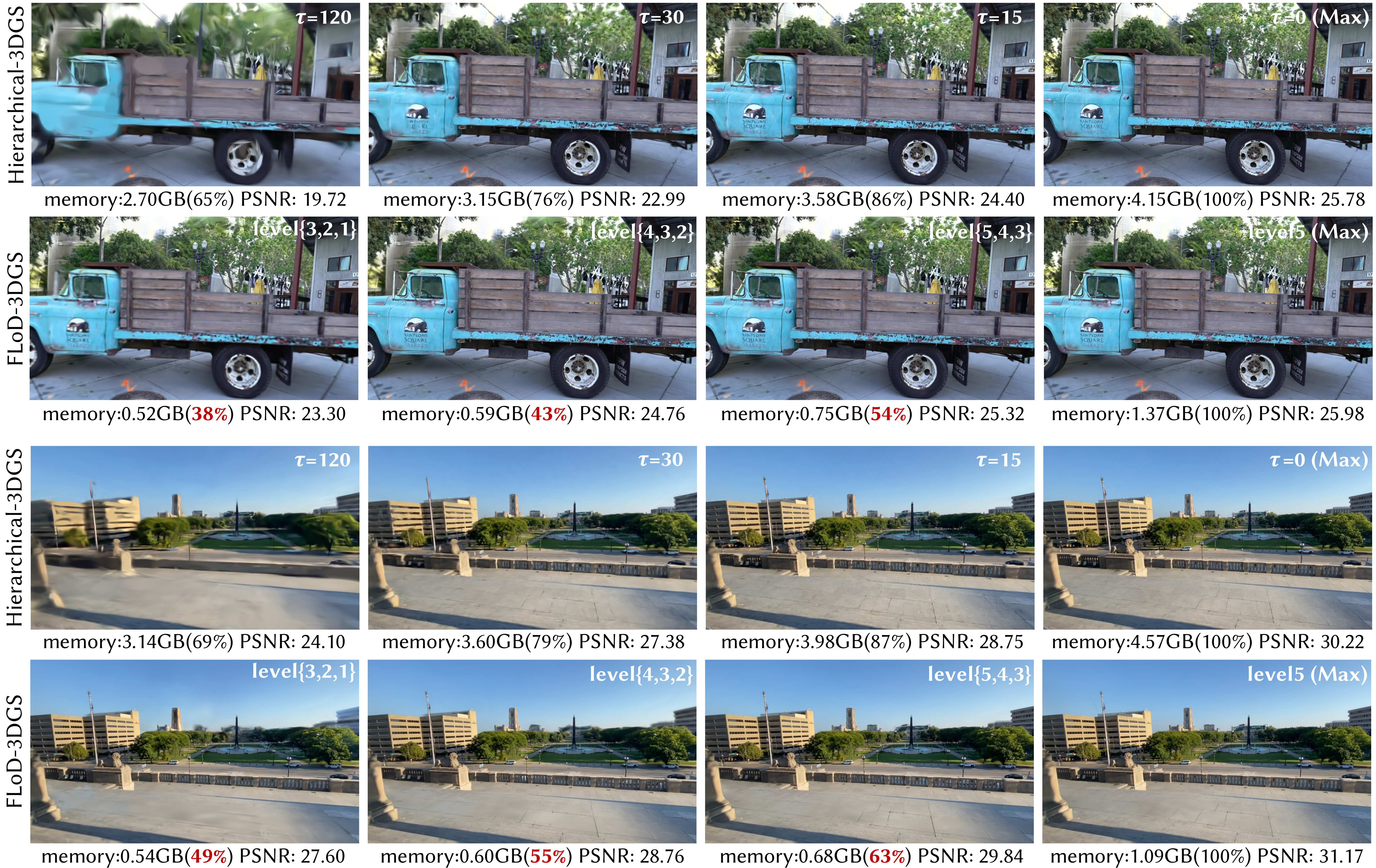}
    \caption{Comparison of the trade-off between memory usage and visual quality in the selective rendering methods of FLoD-3DGS and Hierarchical-3DGS on the Tanks\&Temples and DL3DV-10K datasets. The percentages (\%) next to the memory values indicate how much memory each rendering setting uses compared to the memory required by the setting labeled as "Max" for achieving maximum rendering quality.}
    \label{fig:appendix_selective_more}
\end{figure*}

\subsection{Gaussian Scale Constraint vs Count Constraint}
FLoD controls the level of detail and corresponding memory usage by training Gaussians with explicit 3D scale constraints. Adjusting the 3D scale constraint provides multiple rendering options with different memory requirements, as larger 3D scale constraints result in fewer Gaussians needed for scene reconstruction. 

An alternative method is to create multi-level 3DGS representations by directly limiting the Gaussian count. 
However, limiting the Gaussian count without enforcing scale constraints cannot reconstruct each level’s representation with the level of detail controlled. With only the rendering loss guiding Gaussian optimization and population control, certain local regions may achieve higher detail than others. This regional variation makes visually consistent rendering infeasible when multiple levels are combined for selective rendering, making such rendering option unviable.

In contrast, FLoD’s 3D scale constraints ensure uniform detail within each level. Such uniformity enables visually consistent selective rendering and allows efficient calculation, as \(G_{\text{sel}}\) can be constructed simply by computing the distance \(d_{G^{(l)}}\) of each Gaussian from the camera, as discussed in Section~\ref{sec:selective_rendering}. Furthermore, as discussed in Section~\ref{sec:max_level}, the 3D scale constraints also help preserve scene structure—especially in distant regions. Therefore, limiting the Gaussian count without scale constraints would degrade reconstruction quality.

\begin{figure*}[t]
    \includegraphics[width=\textwidth]{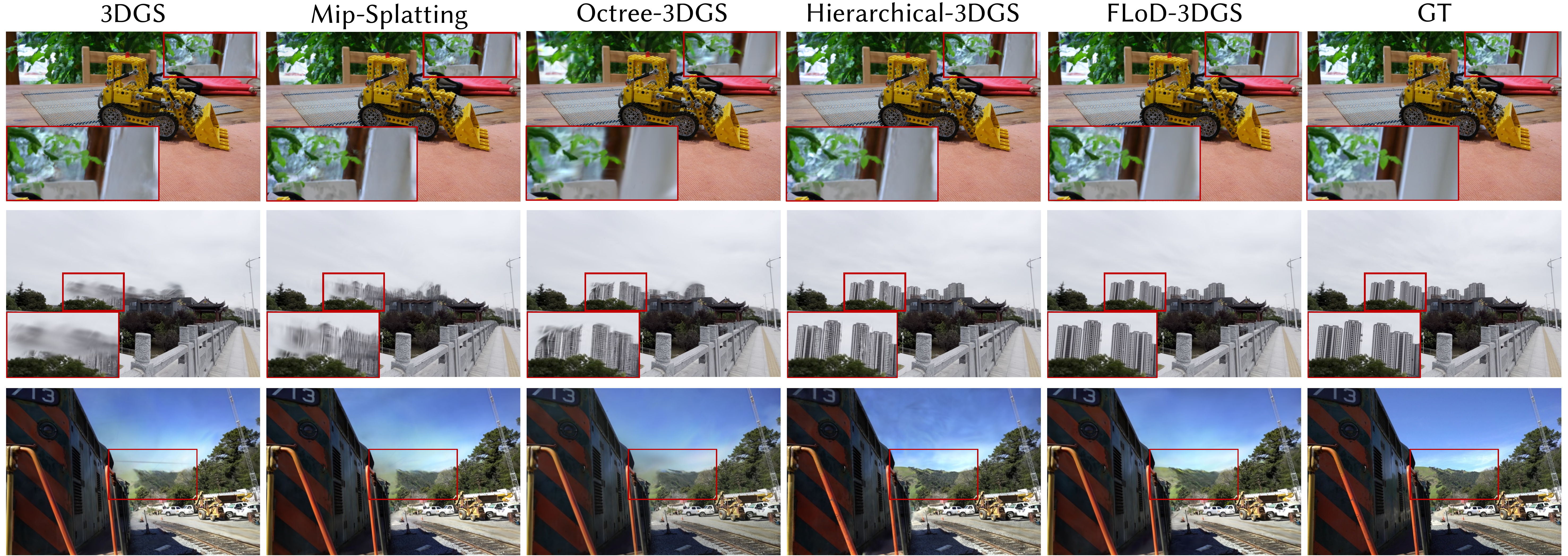}
    \caption{ 
    Qualitative comparison between FLoD-3DGS and baselines on three real-world datasets. 
    The red boxes emphasize the key differences. Please zoom in for a more detailed view.}
    \label{fig:qual_comp}
\end{figure*}

\begin{figure}[t]
    \includegraphics[width=\columnwidth]{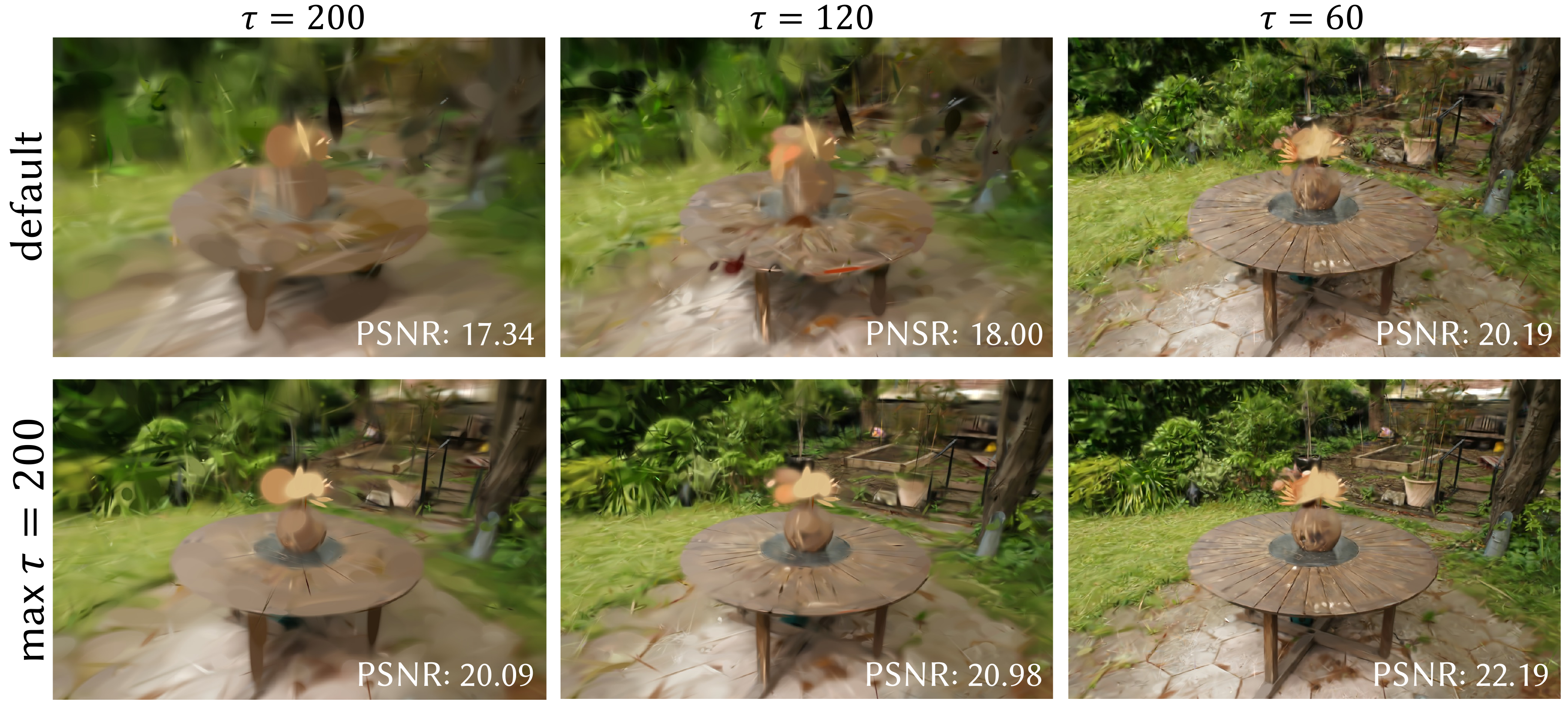}
    \caption{Comparison of Hierarchical-3DGS trained with the default max granularity ($\tau$) and a max $\tau$ of 200. Results show that training with a larger max $\tau$ improves rendering quality for large $\tau$ values.}
    \label{fig:appendix_h3dgs_maxtau}
\end{figure}

\section{Single Level Comparison with Competitors}
\label{sec:appendix_single}

Each level in FLoD has its own independent representation, unlike Octree-GS, where levels are not independent but rather dependent on previous levels. To ensure a fair comparison with Octree-GS in Section \ref{sec:lod_represent}, we respect this dependency. To address any concerns that we may have presented the Octree-GS in a manner advantageous to our approach, we also render results using only the representation of each individual Octree-GS level. These results are shown in the upper row of Figure~\ref{fig:appendix_per_level}. 
As illustrated, Octree-GS automatically assigns higher levels to regions closer to training views and lower levels to more distant regions. This characteristic limits its flexibility compared to FLoD-3DGS, as it cannot render using various subsets of levels.

In contrast, Hierarchical-3DGS automatically renders using nodes across multiple levels based on the target granularity \(\tau\). 
It does not support rendering with nodes from a single level, unlike FLoD-3DGS and Octree-GS. 
For this reason, we do not conduct single-level comparisons for Hierarchical-3DGS in Section \ref{sec:lod_represent}. 
However, to offer additional clarity, we render using only nodes from five selected levels (1, 6, 11, 16, and 22) out of its 22 levels. These results are shown in the lower row of Figure~\ref{fig:appendix_per_level}.

\section{Selective Rendering Comparison}
\label{sec:appendix_selective}

In Section~\ref{sec:selective_rendering_exp}, we compare the memory efficiency of selective rendering between FLoD-3DGS and Hierarchical-3DGS. 
Since the default setting of Hierarchical-3DGS is intended for a maximum target granularity of 15, we extend the maximum target granularity \(\tau_{max}\) to 200 during its hierarchy optimization stage. 
This adjustment ensures a fair comparison with Hierarchical-3DGS across a broader range of rendering settings.
As shown in Figure~\ref{fig:appendix_h3dgs_maxtau}, its default setting results in significantly worse rendering quality for large $\tau$ compared to when the hierarchy optimization stage has been adjusted.

Section~\ref{sec:selective_rendering_exp} presents results for the garden scene from the Mip-NeRF360 dataset. To demonstrate that FLoD-3DGS achieves superior memory efficiency across diverse scenes, we include additional results for the Tanks\&Temples and DL3DV-10K datasets in Figure~\ref{fig:appendix_selective_more}. 
In Hierarchical-3DGS, increasing the target granularity \(\tau\) does not significantly reduce memory usage, even though fewer Gaussians are used for rendering at larger \(\tau\) values. 
This occurs because all Gaussians, across every hierarchy level, are loaded onto the GPU  according to the release code for evaluation.
Consequently, the potential for memory reduction at higher \(\tau\) values is limited.
The results in Figure~\ref{fig:appendix_selective_more} confirm that FLoD-3DGS effectively balances memory usage and visual quality trade-offs through selective rendering across various datasets.

\section{Inconsistency in Selective Rendering}

\begin{figure}[t]
    \includegraphics[width=\columnwidth]{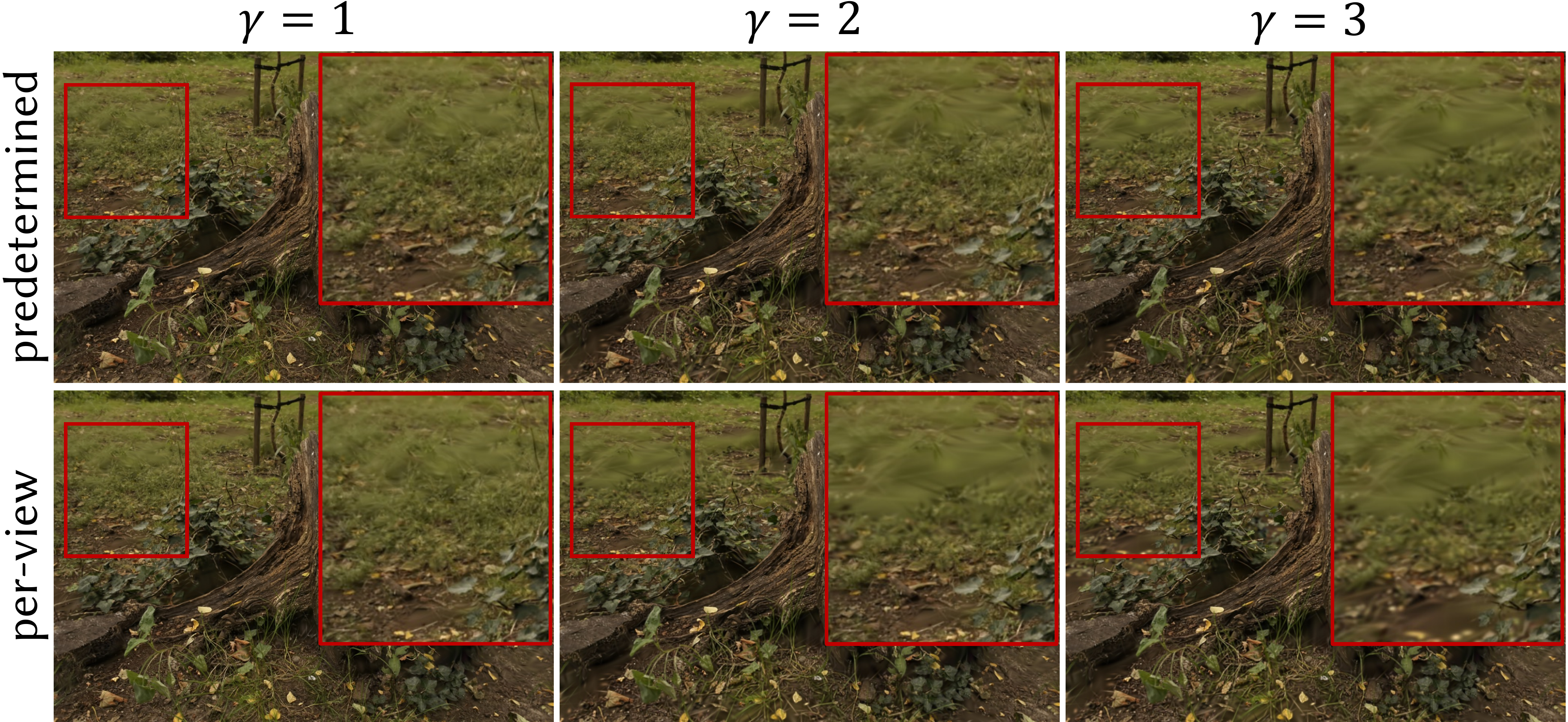}
    \caption{Rendering results of selective rendering using levels 5,4 and 3 with screen size thresholds \(\gamma\) = 1, 2, and 3 for both predetermined and per-view Gaussian set \(\mathbf{G}_\text{sel}\) creation methods on the Mip-NeRF360 dataset. Red boxes emphasize the region where inconsistency is visible for larger \(\gamma\) settings.}
    \label{fig:selective_inconsistency}
\end{figure}

\input{tables/selective_notebook}
\input{tables/appendix_lightcompactgs}

In our selective rendering approach, the transition to a lower level occurs at the distance where the 2D projected 3D scaling constraint for the lower level becomes 1 pixel length, on the default screen size threshold \(\gamma = 1\).  
While lower-level Gaussians can be trained to have large 3D scales - resulting in larger 2D splats - this generally happens when the larger splat aligns well with the training images. 
In such cases, these Gaussians do not receive training signals to shrink or split, and thus retain their large 3D scales. Therefore, inconsistency due to level transitions in selective rendering is unlikely, which is why we did not implement interpolation between successive levels.
On the other hand, increasing the screen size threshold \(\gamma\) beyond 1 can introduce visible inconsistencies in the rendering, as shown in Figure~\ref{fig:selective_inconsistency}.

\section{Qualitative Results of Max-level Rendering}
\label{sec:appendix_maxlevel_rendering}
Section~\ref{sec:max_level} quantitatively demonstrates that FLoD achieves rendering quality comparable to existing models. Figure~\ref{fig:qual_comp} qualitatively shows that FLoD-3DGS reconstructs thin details and distant objects more accurately, or at least comparably, to the baselines. While Hierarchical-3DGS also handles distant objects well, it receives depth information from an external model. In contrast, FLoD-3DGS is trained without extra supervision.

\section{Rendering on Low-cost Device}
\label{sec:appendix_lowcost}

FLoD offers wide range of rendering options through single-level and selective rendering, allowing users to adapt to a wide range of hardware capabilities. 
To demonstrate its effectiveness on low-cost devices, we measure FPS for Mip-NeRF360 scenes on the laptop equipped with an MX250 GPU (2GB VRAM).

As shown in Table~\ref{tab:selective_notebook}, single-level rendering at level 5 causes out-of-memory (OOM) errors in some scenes (e.g., stump). However, using selective rendering with levels 5, 4, and 3, or switching to a lower single level, resolves these errors. Additionally, in some cases (e.g., bonsai), FLoD enables real-time rendering. Thus, FLoD can provide adaptable rendering options even for low-cost devices.

\section{Comparison with compression methods}
LightGaussian~\cite{fan2023lightgaussian} and CompactGS~\cite{lee2023compact} also address memory-related issues, but their primary focus is on creating a single compressed 3DGS with small storage size. In contrast, FLoD constructs multi-level LoD representations to accommodate varying GPU memory capacities during rendering.
Due to this difference in purpose, a direct comparison with FLoD was not included in the main paper.

To demonstrate the efficiency of FLoD-3DGS in GPU memory usage during rendering, we compare PSNR and GPU memory consumption across levels 5, 4, and 3 of FLoD-3DGS and the two baselines.
As shown in Table~\ref{tab:appendix_lightcompactgs}, FLoD-3DGS achieves higher PSNR with comparable GPU memory usage. Furthermore, unlike LightGaussian and CompactGS, FLoD-3DGS supports multiple memory usage settings, indicating its adaptability across a range of GPU settings.

\input{tables/appendix_lightgs}
\section{LightGaussian Compression on FLoD-3DGS}
FLoD-3DGS can store and render specific levels as needed. However, keeping the option of rendering with all levels requires significant storage disk space to accommodate them.
To address this, we integrate LightGaussian's \cite{fan2023lightgaussian} compression method into FLoD-3DGS to reduce storage disk usage. As shown in Table~\ref{tab:appendix_limit_lightgs}, compressing FLoD-3DGS reduces storage disk usage by 93\% and enhances rendering speed.
This compression, however, results in a reduction in reconstruction quality metrics compared to the original FLoD-3DGS, similar to how LightGaussian shows lower reconstruction quality than its baseline model, 3DGS.  Despite this, we demonstrate that FLoD-3DGS can be further optimized to suit devices with constrained storage by incorporating compression techniques.

%% file: tables/algorithm.tex
\begin{algorithm}[t]
\SetAlgoNoLine
\caption{Overall Training Algorithm for FLoD-3DGS \\
$L_{\text{max}}$: maximum level \\
$\lambda, \rho$: 3D scale constraint at level 1, scale factor 
}

$M \gets \text{SfM Points}$ \hfill $\triangleright$ Positions

$S, R, C, A \gets \text{InitAttributes}()$ \hfill $\triangleright$ Scales, Rotations, Colors, Opacities

\For{$l = 1$ ... $L_\text{max}$}{
    \eIf{$l < L_\text{max}$}{
        $s_{\text{min}}^{(l)} \gets \lambda \times \rho^{1-l}$ \hfill $\triangleright$ 3D Scale constraint for current level
    }{
        $s_{\text{min}}^{(l)} \gets 0$ \hfill $\triangleright$ No constraint at maximum level
    }

    $i \gets 0$ \hfill $\triangleright$ Iteration count

    \While{\textnormal{not converged}}{

        $S^{(l)} \gets \text{ApplyScaleConstraint}(S_\text{opt}, s_{\text{min}}^{(l)})$ \hfill $\triangleright$ Eq.4

        $I \gets \text{Rasterize}(M, S^{(l)}, R, C, A)$

        $L \gets \text{Loss}(I, \hat{I})$

        $M, S_\text{opt}, R, C, A \gets \text{Adam}(\nabla L)$ \hfill $\triangleright$ Backpropagation

        \If{$i < \textnormal{DensificationIteration}$}{
            \If{$\textnormal{RefinementIteration}(i, l)$}{
                $\textnormal{Densification}()$
                
                $\textnormal{Pruning}()$
                
                $\textnormal{OverlapPruning}()$ \hfill $\triangleright$ Overlap pruning step
            }
        }

        $i \gets i + 1$
    }

    $\text{SaveClone}(l, M, S^{(l)}, R, C, A)$ \hfill $\triangleright$ Save clones for level $l$

    \If{$l \neq L_{\text{max}}$}{
        $S_\text{opt} \gets \text{AdjustScale}(S^{(l)})$ \hfill $\triangleright$ Adjust scales for level $l+1$
    }
}
\label{tab:algorithm}
\end{algorithm}

%% file: tables/selective_notebook.tex
\begin{table*}[t]
\caption{
Rendering FPS results of FLoD-3DGS on a laptop with MX250 2GB GPU for 7 scenes from the Mip-NeRF360 dataset. 
A "\cmark" on a single level indicates single-level rendering, while a "\cmark" on multiple levels indicates selective rendering. 
"\xmark" represents an OOM error, indicating that rendering FPS could not be measured.
}
\centering
\begin{tabular}{ccccc|ccccc|cccc}
\hline
\multicolumn{5}{c|}{Level} & \multicolumn{5}{c|}{Outdoor} & \multicolumn{4}{c}{Indoor} \\
1 & 2 & 3 & 4 & 5 & $\text{bicycle}$ & $\text{flowers}$ & $\text{garden}$ & $\text{stump}$ & $\text{treehill}$ & $\text{room}$ & $\text{counter}$ & $\text{kitchen}$ & $\text{bonsai}$ \\
\hline
\multicolumn{5}{r|}{\cmark} & {\xmark} & 6.52 & {\xmark} & {\xmark} & 5.77 & 5.54 & 6.00 & 3.99 & 7.48 \\
\multicolumn{5}{r|}{\cmark - \cmark - \cmark} & 5.10 & 8.81 & 6.92 & 8.48 & 8.33 & 6.27 & 6.58 & 4.20 & 8.69 \\
\multicolumn{5}{r|}{{\cmark} \phantom{$/\checkmark$}}  & 7.71 & 10.25 & 7.27 & 10.41 & 9.87 & 8.35 & 8.71 & 5.67 & 9.16 \\
\multicolumn{5}{c|}{\cmark - \cmark - \cmark}  & 8.53 & 11.38 & 7.98 & 13.20 & 11.39 & 8.42 & 8.79 & 5.73 & 9.31 \\
\multicolumn{5}{c|}{\cmark} & 9.21 & 15.00 & 13.54 & 18.19 & 12.97 & 9.67 & 11.65 & 10.44 & 11.68 \\
\multicolumn{5}{l|}{\cmark - \cmark - \cmark} & 9.34 & 15.60 & 13.98 & 20.92 & 13.77 & 9.72 & 11.73 & 10.49 & 11.85 \\
\hline
\end{tabular}
\label{tab:selective_notebook}
\end{table*}


%% file: tables/appendix_lightcompactgs.tex
\begin{table}[t]
\caption{
Comparison of visual quality and memory usage (GB) for FLoD-3DGS, alongside LightGS and CompactGS on Mip-NeRF360(Mip), DL3DV-10K(DL3DV) and Tanks\&Temples(T\&T) datasets.
}
\centering
\setlength{\tabcolsep}{1mm}
\begin{tabular}{l|cc|cc|cc}
\hline
\multicolumn{1}{c|}{} & \multicolumn{2}{c|}{Mip} & \multicolumn{2}{c|}{DL3DV} & \multicolumn{2}{c}{T\&T} \\
& $\text{PSNR}$ & mem. & $\text{PSNR}$ & mem. & $\text{PSNR}$ & mem. \\
\hline
FLoD-3DGS(lv5) & 27.8 & 1.8 & 31.9 & 1.0 & 24.4 & 1.1 \\
FLoD-3DGS(lv4) & 26.6 & 1.2 & 30.7 & 0.6 & 23.8 & 0.6 \\
FLoD-3DGS(lv3) & 24.1 & 0.8 & 28.3 & 0.5 & 21.7 & 0.5 \\
\hline
LightGS & 26.6 & 1.2 & 27.2 & 0.7 & 23.3 & 0.6 \\
CompactGS & 26.8 & 1.1 & 27.8 & 0.5 & 22.8 & 0.8 \\
\hline
\end{tabular}

\label{tab:appendix_lightcompactgs}
\end{table}

%% file: tables/appendix_lightgs.tex
\begin{table}[t]
\caption{
Comparison of Level 5 single-level rendering between FLoD-3DGS and FLoD-3DGS with the LightGaussian compression method applied (denoted as '+LightGS') on the Mip-NeRF360 dataset.
}
\centering
\setlength{\tabcolsep}{4pt}
\begin{tabular}{l|ccccc}
\hline
& $\text{FPS}$ & $\text{Size(MB)}$ & $\text{PSNR}$ & $\text{SSIM}$ & $\text{LPIPS}$ \\
\hline
FLoD-3DGS & 103 & 518 & 27.8 & 0.815 & 0.224 \\
FLoD-3DGS+LightGS & 144 & 31.7 & 27.1 & 0.799 & 0.250 \\
\hline
\end{tabular}

\label{tab:appendix_limit_lightgs}
\end{table}